\documentclass[pdflatex,sn-mathphys,iicol]{sn-jnl}
\usepackage[T1]{fontenc}
\usepackage{anyfontsize}

\usepackage{
graphicx,
amsfonts,
textcomp,
program,
amssymb,
listings,
rotating,
booktabs,
algorithmicx,
mathrsfs,
url,
appendix,
etoolbox,
algpseudocode,
multirow,
algorithm,
amsmath,
hyperref,
manyfoot,
hhline,
pifont}

\usepackage[utf8]{inputenc} 
\usepackage[T1]{fontenc}    
\usepackage{hyperref}       
\usepackage{url}            
\usepackage{booktabs}       
\usepackage{amsfonts}       
\usepackage{nicefrac}       
\usepackage{microtype}      
\usepackage{placeins}
\usepackage{colortbl} 
\usepackage{nicefrac}
\usepackage{dsfont}
\usepackage{marvosym}
\usepackage{fdsymbol}
\usepackage{graphicx}
\usepackage{adjustbox}

\usepackage{overpic}
\usepackage{makecell}
\usepackage{wrapfig}
\usepackage{marvosym}
\usepackage{caption}
\usepackage{colortbl}
\usepackage{xcolor}
\usepackage{nicefrac}
\usepackage{dsfont}
\usepackage{fdsymbol}
\usepackage{amssymb}
\usepackage{amsmath}
\usepackage{balance}
\usepackage{amsmath}
\usepackage{amssymb}
\usepackage{mathtools}
\usepackage{amsthm}

\usepackage[capitalize,noabbrev]{cleveref}

\theoremstyle{plain}

\theoremstyle{definition}

\theoremstyle{remark}

\usepackage{multirow}
\usepackage{pifont}
\newcommand{\best}[1]{\textbf{\textcolor{red}{#1}}}
\newcommand{\secondbest}[1]{\underline{\textcolor{blue}{#1}}}
\usepackage{microtype}
\usepackage{graphicx}
\usepackage{subfigure}
\usepackage{ragged2e}
\definecolor{OursColor}{HTML}{FFFFCC}  
\definecolor{myIDBcolor}{HTML}{FFF5F0}
\definecolor{myCCDBcolor}{HTML}{F5FFF0}  

\newcommand\shline{\specialrule{0.8pt}{0pt}{0pt}}

\begin{document}

\title[Article Title]{VQ-Jarvis: Retrieval-Augmented Video Restoration Agent \\ with Sharp Vision and Fast Thought}









\author[1,2]{\fnm{Xuanyu} \sur{Zhang}}\email{xuanyuzhang21@stu.pku.edu.cn}

\author[1,2]{\fnm{Weiqi} \sur{Li}}\email{liweiqi@stu.pku.edu.cn}

\author[2]{\fnm{Qunliang} \sur{Xing}}\email{xingqunliang@bytedance.com}

\author[2]{\fnm{Jingfen} \sur{Xie}}\email{xiejingfen@bytedance.com}

\author[1,2]{\fnm{Bin} \sur{Chen}}\email{chenbin@stu.pku.edu.cn}

\author[2]{\fnm{Junlin} \sur{Li}}\email{lijunlin.li@bytedance.com}

\author[2]{\fnm{Li} \sur{Zhang}}\email{lizhang.idm@bytedance.com}

\author[1]{\fnm{Jian} \sur{Zhang}}\email{zhangjian.sz@pku.edu.cn}

\author*[2]{\fnm{Shijie} \sur{Zhao}}\email{zhaoshijie.0526@bytedance.com}

\affil[1]{\orgdiv{School of Electronic and
Computer Engineering, Peking University}, 
\orgaddress{\city{Shenzhen}, \country{China}}}
\affil[2]{\orgdiv{ByteDance Inc.}, 
\orgaddress{\city{Shenzhen}, \country{China}}}


\abstract{Video restoration in real-world scenarios is challenged by heterogeneous degradations, where static architectures and fixed inference pipelines often fail to generalize. Recent agent-based approaches offer dynamic decision making, yet existing video restoration agents remain limited by insufficient quality perception and inefficient search strategies. We propose VQ-Jarvis, a retrieval-augmented, all-in-one intelligent video restoration agent with sharper vision and faster thought. VQ-Jarvis is designed to accurately perceive degradations and subtle differences among paired restoration results, while efficiently discovering optimal restoration trajectories. To enable sharp vision, we construct VSR-Compare, the first large-scale video paired enhancement dataset with 20K comparison pairs covering 7 degradation types, 11 enhancement operators, and diverse content domains. Based on this dataset, we train a multiple operator judge model and a degradation perception model to guide agent decisions. To achieve fast thought, we introduce a hierarchical operator scheduling strategy that adapts to video difficulty: for easy cases, optimal restoration trajectories are retrieved in a one-step manner from a retrieval-augmented generation (RAG) library; for harder cases, a step-by-step greedy search is performed to balance efficiency and accuracy. Extensive experiments demonstrate that VQ-Jarvis consistently outperforms existing methods on complex degraded videos.}


\maketitle

\section{Introduction}

Video restoration aims to recover high-quality, perceptually faithful, and temporally consistent video content from degraded observations. As a fundamental problem in computer vision, it plays a critical role in a wide range of real-world applications, including video streaming~\cite{wang2023reparo, zhuang2025flashvsr}, surveillance, autonomous driving~\cite{lin2025jarvisir}, content creation~\cite{yang2024cogvideox}, and downstream visual understanding~\cite{li2025q}. Unlike image restoration, video restoration must simultaneously address spatial fidelity and temporal coherence, making it inherently more challenging.

In real-world scenarios, video degradation is rarely simple or isolated~\cite{liu2025moa, potlapalli2023promptir}. Videos captured and transmitted in uncontrolled environments often suffer from complex and heterogeneous degradations, such as noise, compression artifacts~\cite{li2025uare}, blur, low resolution~\citep{wang2026gendr, chen2026improved}, and low-light conditions~\citep{yang2025difflle, peng2022lve}. These degradations are typically mixed, temporally varying, and strongly coupled across frames, leading to non-linear error accumulation during restoration. As a result, restoration methods designed under simplified or single-degradation assumptions often fail to generalize reliably to real-world videos.

Existing video restoration methods generally follow either task-specific designs~\cite{chan2022basicvsr++}, where a dedicated model is trained for a particular degradation, or all-in-one paradigms~\cite{potlapalli2023promptir, liang2024vrt} that attempt to handle multiple degradations within a single network. While both directions have achieved progress, they fundamentally rely on static architectures and fixed inference pipelines, which limits their adaptability to the diversity and unpredictability of real-world degradation patterns. In contrast, recent advances in agent-based systems have demonstrated strong capabilities in handling complex, open-ended tasks through dynamic reasoning, tool selection, and feedback-driven decision making. Agent frameworks have been successfully applied to code generation~\cite{li2025codetree}, multimodal reasoning, scientific discovery, and image restoration~\cite{liu2025moa, zhu2024intelligent, jiang2025multi}, suggesting a broader trend toward modular and adaptive intelligence systems. 

Video restoration is well-suited to the agentic system. Different restoration operators exhibit complementary strengths across data domains (e.g., AIGC content, human faces, natural scenes) and degradation types. No single operator performs optimally across all conditions, and the effectiveness of a method often depends on both the degradation composition and the execution order. An agent-based framework enables the system to analyze degradation characteristics, dynamically select appropriate restoration tools, and organize them into effective restoration trajectories, rather than relying on manually designed or fixed pipelines.

Despite emerging interest, current agent-based approaches~\cite{liu2025moa, zhou2025q,zuo20254kagent} for video restoration remain limited in several key aspects. \textbf{First}, systematic studies on video restoration agents are still scarce.
Compared to image restoration, video restoration introduces a significantly larger decision space due to temporal dependencies and degradation propagation across frames. Most existing agent-based methods are either image-centric or only loosely extended to video settings, without fully addressing the unique challenges of temporal consistency and compound degradations. \textbf{Second}, existing agent systems lack awareness of performance boundaries of different restoration operators.
Most agent frameworks rely on generic, model-based quality metrics, such as CLIPIQA~\cite{wang2023exploring} or ManIQA~\cite{yang2022maniqa}, to evaluate restoration results. However, these metrics are primarily trained on natural image distributions and are often insensitive to paired enhancement scenarios, where competing restoration results share the same source content and differ only subtly in perceptual quality. As a consequence, agents may fail to accurately judge which restoration truly improves quality, or may select the sub-optimal methods. \textbf{Third}, current agent search pipelines~\cite{zhu2024intelligent, zhou2025q} are inefficient and heavily dependent on heuristic trial-and-error. Most existing approaches adopt greedy, step-by-step exploration with frequent rollback and re-evaluation, leading to high computational cost and long inference time. In practice, however, videos with similar degradation patterns often share highly consistent optimal restoration strategies. This suggests that blindly searching the restoration space is unnecessary. 

\begin{figure*}[t!]
	\centering
    \includegraphics[width=1.0\linewidth]{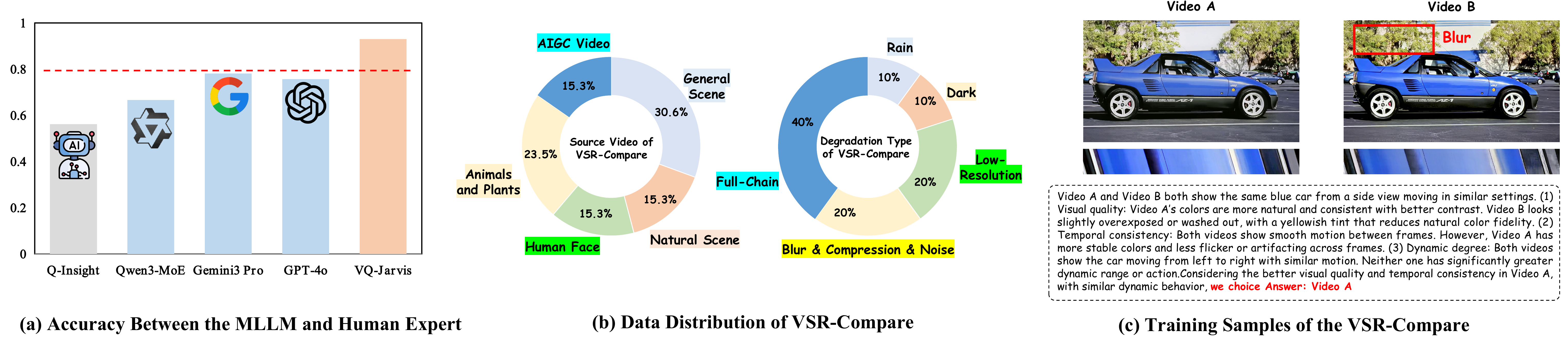}
	\caption{Overview of the challenges in existing enhancement comparison and the construction of VSR-Compare.
(a) Existing quality understanding models (e.g., Q-Insight) fail to reliably capture subtle differences between enhancement results (Acc.~$<$~60\%).
(b) Data domain and degradation distributions of our VSR-Compare.
(c) Training sample from VSR-Compare.}
	\label{intro}
\end{figure*} 

To address the above challenges, we propose an all-in-one intelligent video restoration agent, termed VQ-Jarvis. Our core objectives are twofold: \textbf{(1) ``Sharp vision''}: enabling the agent to more accurately perceive video degradations and identify subtle differences among paired enhancement results, thereby making reliable preference decisions; \textbf{(2) ``Fast thought''}: equipping the agent to discover optimal restoration trajectories more efficiently. To achieve the first objective, we adopt a human machine collaborative annotation paradigm and construct the first large-scale video paired enhancement dataset, VSR-Compare, comprising approximately 20K comparison pairs. The dataset includes both single-degradation enhancement comparisons and full-chain enhancement comparisons. Based on this dataset, we train the preference comparison model and the degradation perception model of VQ-Jarvis. To achieve the second objective, we design a hierarchical operator scheduling strategy. Leveraging prior comparison results, we build a retrieval-augmented generation (RAG)~\cite{li2026qwen3} library containing restoration trajectories for 1K videos. For videos with relatively low restoration difficulty, we employ a RAG-based strategy that directly retrieves and applies the restoration trajectory of the most similar video. For more challenging cases, we adopt a step-by-step greedy search algorithm to balance restoration accuracy and efficiency.

Our contributions can be summarized as follows.

\ding{113}~(1) We propose VQ-Jarvis, a retrieval-augmented, all-in-one video restoration agent. It is characterized by a strong sensitivity to subtle differences in both degraded and enhanced videos, an enhanced ability to efficiently search for optimal restoration trajectories, and the incorporation of prior knowledge over restoration operators.
\vspace{2pt}

\ding{113}~(2) We construct the first video paired enhancement dataset, covering 7 degradation types, 11 enhancement operators, and diverse domains. By leveraging cross-validated preference scores from multiple MLLM experts, complemented with verification and refinement by human experts, we curate 20K enhancement comparison pairs. This dataset effectively improves the model’s ability to distinguish subtle differences among restoration results.
\vspace{2pt}

\ding{113}~(3) We propose a hierarchical operator scheduling strategy. It adaptively selects the strategy based on the difficulty of the input video, either retrieving the optimal restoration trajectory in a one-step manner from the RAG library, or performing a step-by-step search over multiple restoration operators guided by our carefully trained comparison model.
\vspace{2pt}

\ding{113}~(4) Extensive experiments demonstrate that our approach significantly outperforms existing methods on real-world super resolution and multiple degradation restoration tasks, validating both its effectiveness and generalization ability.
\vspace{2pt}

\section{Related Work}
\subsection{Real-World Video Super-Resolution}
Real-world video super-resolution (Real-VSR) \citep{tao2017detail,nah2019ntire} studies how to reconstruct high-resolution (HR) videos from low-resolution (LR) observations affected by unknown, real-world degradations. Training data are typically obtained either by capturing aligned LR/HR pairs with different camera configurations \citep{yang2021real,wang2023benchmark}, or by generating LR inputs from HR videos using composite degradation pipelines (e.g., blur/noise/resizing/compression) with randomized orders and strengths \citep{wang2021real,zhang2021designing,chan2022investigating}. 

Based on these datasets, many Real-VSR models have been explored \citep{shi2022rethinking}. Representative reconstruction-based approaches, such as BasicVSR \citep{chan2021basicvsr,chan2022basicvsr++}, EDVR \citep{wang2019edvr}, and RVRT \citep{liang2024vrt,liang2022recurrent}, are effective at distortion suppression but may become overly smooth under heavy degradations. More recently, diffusion-based methods have improved perceptual realism for Real-VSR. For example, Upscale-A-Video \citep{zhou2024upscale}, VEnhancer \citep{he2024venhancer}, and STAR \citep{xie2025star} inject LR-conditioned transformation/control modules into diffusion backbones. MGLD-VSR \citep{yang2024motion} enhances temporal coherence with motion guidance. SeedVR \citep{wang2025seedvr} adopts shifted-window DiTs to better handle varying spatial resolutions. To reduce sampling latency, recent work explores one-step diffusion. For example, SeedVR2 further \citep{wang2025seedvr2} progressively distills a pretrained SeedVR \citep{wang2025seedvr} into one sampling step and enhances it through adversarial post-training. DOVE \citep{chen2025dove} adapts pretrained CogVideoX networks \citep{yang2024cogvideox} to Real-VSR by fine-tuning them on a curated high-quality video dataset. FlashVSR \citep{zhuang2025flashvsr} proposes a one-step streaming diffusion design that enables real-time inference and scales to ultra-high resolutions.

\begin{figure*}[t!]
	\centering
    \includegraphics[width=1.0\linewidth]{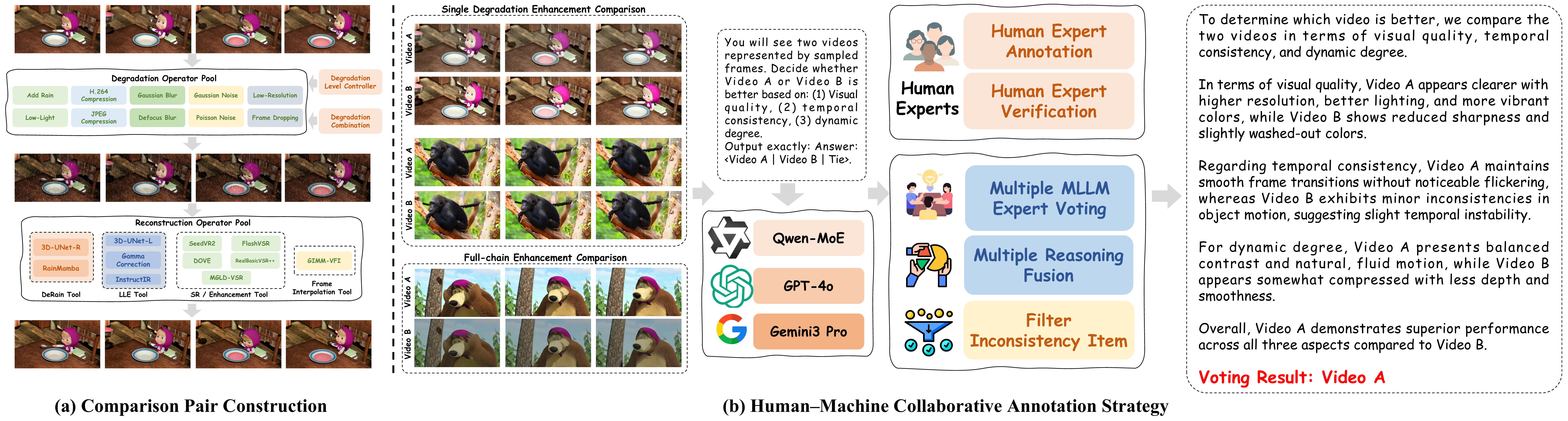}
	\caption{Overview of VSR-Compare pipeline, including degradation-aware pair construction, human machine collaborative annotation, and the data distribution of VSR-Compare. The outputs of multiple MLLMs are first filtered to remove inconsistent selections, and their reasoning results are fused; together with additional human expert annotation and verification, they jointly form the comparison pairs.
}
	\label{benchmark}
\end{figure*}

\subsection{Agent-Based Restoration}
Recent image and video restoration research has moved beyond end-to-end models toward agent-based restoration frameworks, where restoration is formulated as a sequential decision-making process. In this paradigm, large multi-modal language models act as reasoning agents that analyze degradations, plan restoration steps, and invoke task-specific restoration tools in an adaptive manner. For instance, RestoreAgent shows that degradation diagnosis and dynamic tool composition can improve robustness over conventional all-in-one models~\cite{chen2024restoreagent}. Building on this idea, Q-Agent introduces quality-driven chain-of-thought reasoning, enabling the agent to assess intermediate results and refine restoration strategies to avoid over- or under-processing~\cite{zhou2025q}. Other agentic systems emphasize diagnostic reasoning and closed-loop control for complex restoration scenarios~\cite{zhu2024intelligent}. Beyond single-agent formulations, multi-agent image restoration explore collaborative agent designs, showing that task decomposition and inter-agent coordination further enhance flexibility and generalization across diverse degradations~\cite{li2025hybrid, jiang2025multi}. Agent-based restoration has also been applied to practical settings, such as JarvisIR for autonomous driving perception, where restoration decisions are aligned with downstream tasks~\cite{lin2025jarvisir}. Extensions to high-resolution and video restoration, including 4KAgent and MoA-VR, further indicate the scalability of agent-based designs to more challenging spatial and temporal domains~\cite{zuo20254kagent, liu2025moa}. Despite these advances, agent-based video restoration methods remain relatively underexplored. Existing video restoration agents are still constrained by inaccurate quality perception, outdated operator pools, and low search efficiency.

\section{VSR-Compare Benchmark Construction}

\subsection{Motivation and Challenges}
The ability to perform \emph{paired enhancement comparison} is a fundamental capability for video restoration agent. It directly determines whether the agent can accurately perceive quality differences among multiple operators and make correct decisions. However, we observe that existing visual quality understanding models (e.g., Q-Insight~\cite{li2025q}, RALI~\citep{zhao2025reasoning}, CLIPIQA~\cite{wang2023exploring}) are inadequate for this purpose. Most of these models are trained primarily on degraded data and focus on estimating degradation degree of videos. Thus, they are relatively insensitive to subtle yet critical quality differences among restored videos. As shown in Fig.~\ref{intro}~(a), we randomly sample 100 pairs of reconstruction results produced by multiple degradation operators for comparison using Q-Insight, and observe that its agreement with human experts is only 56\%. Meanwhile, constructing high-quality datasets for video paired enhancement comparison remains highly challenging. Existing paired comparison datasets are extremely limited in both scale and diversity. While prior works such as DiffIQA~\cite{chen2025toward} perform comparisons across multiple image super-resolution methods, they are neither exposed to nor suitable for emerging video restoration techniques. More importantly, real-world video restoration often involves \emph{high-order and composite degradations}, where effective enhancement requires a chain of multiple restoration operators rather than a single model. This significantly increases the complexity of dataset construction. Thus, relying solely on human annotators to judge the quality of entire restoration chains becomes prohibitively expensive and cognitively demanding.

Despite these challenges, we identify \textbf{two key observations}. First, recent advances in general MLLMs enable them to perform paired enhancement comparison with a surprisingly high degree of consistency with human judgments~\cite{zheng2024lm4lv,li2025investigate}. As shown in Fig.~\ref{intro}(a), we observe that general MLLMs such as Gemini3 Pro and GPT-4o, achieve preference alignment accuracies of 78\% and 76\%, respectively, compared with human annotations. This suggests that MLLMs can serve as effective auxiliary annotators for large-scale comparison data. Second, modern video restoration operators~\cite{wang2025seedvr2} are increasingly unified and comprehensive. They tend to incorporate multiple restoration capabilities, such as deblurring and denoising, into a single super-resolution model. This trend substantially reduces the combinatorial complexity of operator selection and dataset construction.

\subsection{Construction of Comparison Pairs}
\textbf{Source video selection.}  
To ensure high visual quality while covering diverse data domains, we curate a heterogeneous set of source videos. Specifically, as shown in Fig.~\ref{intro}, we collect over 2,000 high-quality videos (720p~/~1080p), covering diverse domains including general scenes, animation~\cite{chen2025dove}, human faces~\cite{zhou2024upscale}, animals and plants~\cite{zhou2024upscale}, natural landscapes~\cite{wang2021real}, and AI-generated video. The dataset is split into train and test sets with an approximate ratio of 9:1.

\textbf{Degradation modeling and comparison settings.}  
We consider 7 common real-world degradation types, including low-light, rain, blur, noise, compression, low resolution, and frame dropping. To reduce annotation complexity and facilitate subsequent agent-based usage, we design two complementary comparison scenarios: \emph{single-degradation enhancement comparison} and \emph{full-chain enhancement comparison}. Single-degradation comparison aims to enable the model to perceive the effect of each degradation-specific operator at every restoration step across different videos and degradation levels. Full-chain comparison refers to sequentially applying low-light, rain, one of blur/noise/compression, and low resolution to the source video and compare the final results. More details are reported in the \textbf{Appendix}~\ref{sec:degrade}. Our degradation pipeline is adapted from these two prior works~\cite{zhu2024intelligent, chen2025dove}.

\begin{figure*}[t!]
	\centering
    \includegraphics[width=1.0\linewidth]{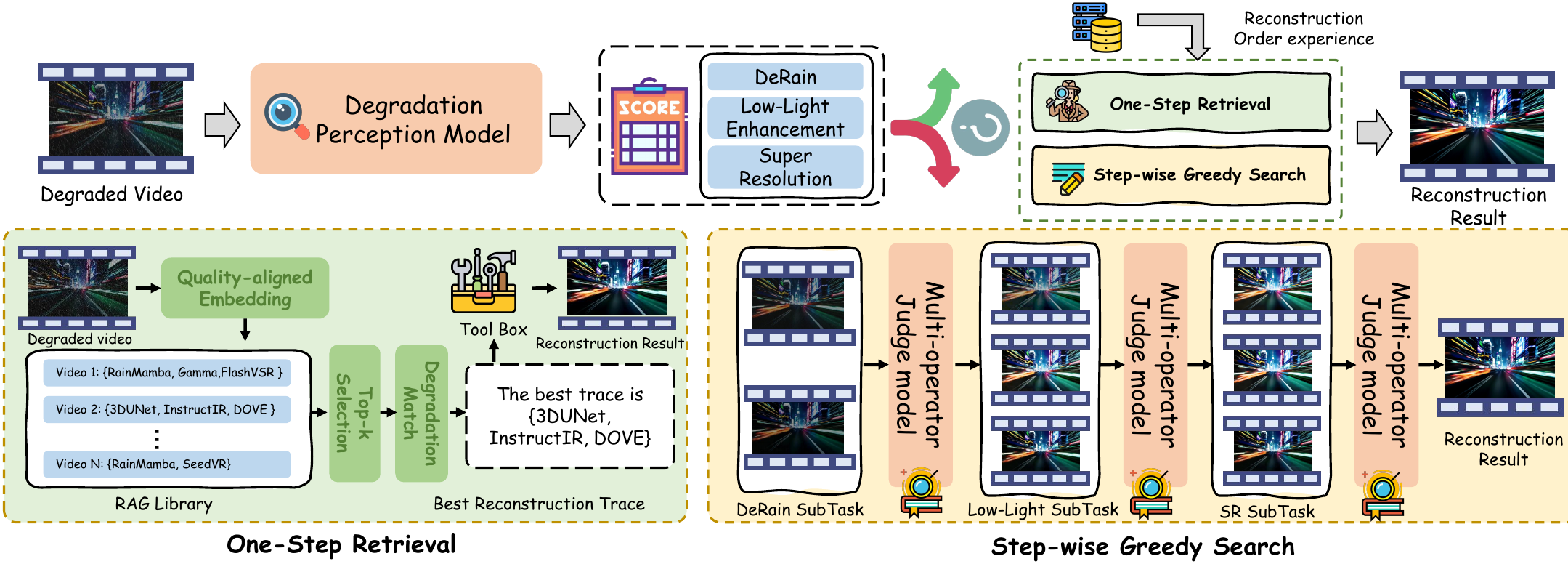}
	\caption{Overview of the proposed VQ-Jarvis framework. Given a degraded input video, a degradation perception model first estimates the video quality score and degradation attributes. Based on the predicted score, the agent adaptively selects between two restoration strategies: one-step retrieval, which retrieves an optimal restoration trajectory from a quality-aligned RAG library using prior reconstruction experience, and step-wise greedy search, which sequentially applies and compares multiple restoration operators across sub-tasks.}
	\label{framework}
\end{figure*}

\textbf{Selection of restoration operators.}
Given the limited availability of dedicated methods for certain degradations (e.g., deraining and low-light enhancement) and their insufficient suitability for our target scenes, we additionally design and train in-house 3D-UNet-based models to enrich the operator pool. 
In addition, a key characteristic of our operator pool is the inclusion of recent, comprehensive real-world video super-resolution operators, such as SeedVR2~\cite{wang2025seedvr2}, FlashVSR~\cite{zhuang2025flashvsr}. 
Benefiting from the strong priors learned by diffusion models, these operators are capable of handling multi-order degradations, such as blur, compression, and noise, in a unified manner. Details of our operators are provided in the \textbf{Appendix}~\ref{sec:operators}.


\textbf{Human machine collaborative annotation strategy.} To reduce the cost of human annotation, we construct 30,000 pairwise comparison samples of enhanced video results. We use Gemini-3 Pro, Qwen-MoE-30B-A3B, and GPT-4o as judge experts, providing them with structured prompts and asking them to output one of the decisions: \texttt{Video~A}, \texttt{Video~B}, \texttt{Tie}.
Subsequently, as shown in Fig.~\ref{benchmark}~(b), we filter out inconsistent items via multi-MLLM expert voting, and fuse their reasoning outputs using Qwen3 to obtain unified reasoning texts together with reliable preference labels, resulting in a total of 18,000 pairs. To further enrich the preference annotations, we recruit 10 human experts and develop a dedicated annotation interface to conduct additional manual annotations and perform random sampling to verify the quality of machine annotations, contributing an additional 2,000 human labeled preference samples. These high-quality paired comparison dataset effectively support the training of our multi-operator judge model. As reported in Fig.~\ref{intro}~(a), the judge model of our VQ-Jarvis can reach 93\% accuracy with human annotators, which far surpasses other MLLMs. 


\section{Methodology}

\subsection{Overall Framework of VQ-Jarvis}
We first formalize multi-degradation video restoration as a combinatorial optimization issue. Given a degraded video $\mathbf{V}_{\text{deg}}$ composed of $K$ degradation types, 
we associate each degradation with a dedicated operator set 
$\mathcal{E}_k = \{ e_{k,1}, \dots, e_{k,N} \}$.
A restoration trajectory is defined as an ordered operator composition.
{\setlength\abovedisplayskip{0.1cm}
\setlength\belowdisplayskip{0.1cm}
\begin{equation}
\mathcal{T} 
= \big( e_{1,i_1},\, e_{2,i_2},\, \dots,\, e_{K,i_K} \big),
\quad e_{k,i_k} \in \mathcal{E}_k,
\end{equation}}where each operator $e_{k,i_k}$ is to address the $k$-th degradation. Our objective is to maximize the perceptual quality of the reconstructed video $\mathbf{V}_{\text{recon}}$:
{\setlength\abovedisplayskip{0.1cm}
\setlength\belowdisplayskip{0.1cm}
\begin{equation}
\mathcal{T}^* = \arg\max_{\mathcal{T}} \; \mathcal{Q}\big( \mathcal{T}(\mathbf{V}_{\text{deg}}) \big),
\end{equation}}where $\mathcal{Q}(\cdot)$ denotes a perceptual quality function defined over restored videos. The optimal restoration trajectory $\mathcal{T}^*$ depends on both the degradation attributes and the intermediate restoration states, resulting in a large and non-trivial search space over operator combinations.

As shown in Fig.~\ref{framework}, VQ-Jarvis addresses this challenge by integrating video quality perception, preference-based comparison, and hierarchical operator scheduling. 
Given $\mathbf{V}_{\text{deg}}$, VQ-Jarvis first uses a degradation perception model $\mathcal{H}_{\text{deg}}$ to estimate a scalar quality score $s_{\text{pred}} \in \mathbb{R}$ and a degradation attribute vector $\mathbf{d}_{\text{pred}}$. 
The predicted quality score serves as an indicator of restoration difficulty and decides the search strategy. 
For videos with $s_{\text{pred}} < \tau$, VQ-Jarvis adopts a step-wise greedy search strategy, progressively selecting the optimal operator for each sub-task based on a multi-operator judge model $\mathcal{H}_{\text{judge}}$. 
Otherwise, a one-step retrieval-based strategy is employed, where the agent directly retrieves a suitable restoration trajectory from a retrieval-augmented generation (RAG) library.
After scheduling the restoration trajectory, the selected operators are executed sequentially to produce the final reconstructed video $\mathbf{V}_{\text{recon}}$.

\begin{table*}[t]
\centering
\caption{Quantitative comparison on UDM10 and YouHQ40 between our VQ-Jarvis and other competitive methods. Overall, our method outperforms all single-operator approaches and surpasses methods that rely on traditional metric DOVER for selection. Best and second-best results are highlighted in \best{red} and \secondbest{blue}, respectively.}
\renewcommand{\arraystretch}{1.15}
\resizebox{\textwidth}{!}{
\begin{tabular}{c l cccccccc}
\toprule
Dataset & Method 
& LPIPS~$\downarrow$ 
& DISTS~$\downarrow$ 
& CLIPIQA~$\uparrow$ 
& MUSIQ~$\uparrow$ 
& MANIQA~$\uparrow$ 
& DOVER~$\uparrow$ 
& Ewarp~$\downarrow$ 
& VQ-Insight~$\uparrow$ \\

\midrule
\multirow{6}{*}{UDM10}

& SeedVR2~\cite{wang2025seedvr2}     
& \best{0.265} & \best{0.153} & 0.347 & 50.09 & 0.523 & 0.327 & 2.56 &50.20 \\

& MGLD-VSR~\cite{yang2024motion}   
& 0.310 & 0.191 & 0.464 & 58.12 & 0.556 & 0.326 & 3.31 & 53.40 \\

& DOVE~\cite{chen2025dove}       
& \best{0.265} & \secondbest{0.174} & \secondbest{0.537} & 60.64 & 0.510 & 0.477 & \best{2.22} & 57.40 \\

& FlashVSR~\cite{zhuang2025flashvsr}  
& 0.288 & 0.175 & 0.521 & \secondbest{65.90} & \best{0.612} & 0.468 & 2.83 & \secondbest{59.60} \\

& DOVER-guide~\cite{wu2023exploring}
& \secondbest{0.276} & 0.175 & 0.503 & 64.27 & 0.548 & \best{0.611} & 3.70 &58.80 \\

& Ours       
& 0.288 & 0.178 & \best{0.538} & \best{66.88} & \secondbest{0.595} & \secondbest{0.479} & \secondbest{2.25} & \best{60.20} \\

\midrule
\multirow{6}{*}{YouHQ40}

& SeedVR2~\cite{wang2025seedvr2}    
& \best{0.284} & \best{0.160} & 0.416 & 61.47 & 0.621 & 0.501 & 5.27 & 57.30 \\

& MGLD-VSR~\cite{yang2024motion}   
& 0.380 & 0.220 & 0.434 & 61.97 & 0.586 & 0.432 & 4.41 & 55.30 \\

& DOVE~\cite{chen2025dove}       
& 0.332 & 0.194 & 0.513 & 59.64 & 0.505 & 0.566 & \best{2.77} & 57.95\\

& FlashVSR~\cite{zhuang2025flashvsr}   
& 0.320 & 0.176 & \secondbest{0.538} & \secondbest{66.41} & \best{0.658} & 0.562 & 4.36 & \secondbest{63.15}\\

& DOVER-guide~\cite{wu2023exploring}
& 0.327 & 0.183 & 0.503 & 64.27 & 0.579 & \best{0.611} & 4.70 & 59.92 \\

& Ours       
& \secondbest{0.312} & \secondbest{0.175} & \best{0.542} & \best{67.79} & \secondbest{0.629} & \secondbest{0.572} & \secondbest{4.12}  & \best{63.50} \\

\bottomrule
\end{tabular}}
\label{realsr}
\end{table*}

\subsection{Degradation Perception and Multi-operator Judge}

VQ-Jarvis relies on two core models: a degradation perception and scoring model and a preference comparison model. Both models are trained using Group Relative Policy Optimization (GRPO)~\cite{guo2025deepseek}, enabling effective learning from preference-based supervision. The degradation perception model is trained via the combination of the quality scoring reward, degradation recognition reward and format reward.

\textbf{Quality Scoring Reward.} To balance reward sparsity and task learning difficulty, we adopt a \emph{hybrid continuous-discrete reward}. Given a predicted quality score $s_{\text{pred}}$ and the ground-truth score $s_{\text{gt}}$, we first define a continuous reward based on relative error:

\begin{equation}
\mathcal{R}_{\text{cont}} = 1 - 
\text{clip}\!\left(
\frac{\|s_{\text{pred}} - s_{\text{gt}}\|}{\|s_{\text{gt}}\| + \delta},
\, 0, \, 1
\right),
\label{eq:score}
\end{equation}

where $\delta$~=~1e-6. We further define a binary reward that encourages strict accuracy within an absolute tolerance:
\begin{equation}
\mathcal{R}_{\text{bin}} = 
\mathbb{I}\!\left(\|s_{\text{pred}} - s_{\text{gt}}\| < \tau_{\text{abs}}\right),
\end{equation}
where $\tau_{\text{abs}}$~=~0.2. $\mathbb{I}(\cdot)$ is the indicator function. The final quality scoring reward is computed as a weighted sum:
\begin{equation}
\mathcal{R}_{\text{score}} = 
w_{\text{cont}} \, \mathcal{R}_{\text{cont}} 
+ 
w_{\text{bin}} \, \mathcal{R}_{\text{bin}},
\end{equation}
where $w_{\text{cont}}$ and $w_{\text{bin}}$ are respectively set to 0.6 and 0.4.

\textbf{Degradation Recognition Reward.} The degradation recognition task is formulated as a multi-label classification problem. We prompt the MLLM to output a one-hot prediction for each predefined degradation category, where 1 indicates the presence of the corresponding degradation and 0 indicates its absence. Let $\mathbf{d}_{\text{pred}}$ denote the predicted degradation labels and $\mathbf{d}_{\text{gt}}$ the ground-truth labels. The corresponding reward is defined as.
\begin{equation}
\mathcal{R}_{\text{deg}} = \mathbb{I}(\mathbf{d}_{\text{pred}} = \mathbf{d}_{\text{gt}}),
\end{equation}

\textbf{Preference Comparison Reward.} We train the multi-operator judge model via preference comparison and format rewards. For paired comparison, the model is given multiple restored candidates $\{\mathbf{V}_i\}_{i=1}^K$ originating from the same degraded video and is required to select the best one. If the model selects candidate $\mathbf{V}_j$ and the ground-truth preferred candidate is $\mathbf{V}_*$, the reward is defined as.
\begin{equation}
\mathcal{R}_{\text{cmp}} = \mathbb{I}(j = *).
\label{eq:comp}
\end{equation}

\begin{table*}[t]
\centering
\caption{Quantitative comparison under 3 multi-degradation settings (Group1, Group2, Group3) on our constructed benchmark.}
\renewcommand{\arraystretch}{1.15}
\resizebox{\textwidth}{!}{
\begin{tabular}{c l ccccccccc}
\toprule
Group & Method & PSNR~$\uparrow$ & SSIM~$\uparrow$ & LPIPS~$\downarrow$ & DISTS~$\downarrow$ & MANIQA~$\uparrow$ & CLIPIQA~$\uparrow$ & MUSIQ~$\uparrow$ & Ewarp~$\downarrow$ & DOVER~~$\uparrow$ \\
\midrule

\multirow{5}{*}{\centering Group1}
& VRT~\cite{liang2024vrt} 
& 15.71 & 0.471 & 0.551 & 0.281 & 0.386 & 0.165 & 33.84 & 15.67 & 0.159 \\
& X-Restormer~\cite{chen2024comparative}
& 15.45 & 0.585 & 0.469 & 0.266 & 0.354 & 0.162 & 32.28 & \best{5.75} & 0.168 \\
& Random Choice
& 16.92 & 0.552 & 0.412 & 0.214 & 0.421 & 0.301 & 52.47 & 9.86 & 0.342 \\
& Qwen3-MoE~\cite{bai2025qwen3vl}
& \secondbest{17.83} & \secondbest{0.568} & \best{0.338} & \secondbest{0.187} & \secondbest{0.496} & \secondbest{0.358} & \secondbest{58.91} & 8.12 & \secondbest{0.417} \\
& Ours
& \best{18.70} & \best{0.584} & \secondbest{0.340} & \best{0.163} & \best{0.582} & \best{0.416} & \best{64.35} & \secondbest{7.01} & \best{0.507} \\

\midrule
\multirow{5}{*}{\centering Group2}
& VRT~\cite{liang2024vrt} 
& \secondbest{22.61} & 0.686 & 0.455 & 0.201 & 0.484 & 0.224 & 41.81 & 22.58 & 0.389 \\
& X-Restormer~\cite{chen2024comparative}
& \best{26.07} & \best{0.766} & 0.287 & 0.147 & 0.410 & 0.200 & 39.66 & \best{4.85} & 0.384 \\
& Random Choice
& 21.38 & 0.691 & 0.263 & 0.132 & 0.512 & 0.336 & 59.24 & 7.96 & 0.451 \\
& Qwen3-MoE~\cite{bai2025qwen3vl}
& 22.01 & 0.707 & \secondbest{0.241} & \secondbest{0.118} & \secondbest{0.548} & \secondbest{0.378} & \secondbest{62.88} & 6.41 & \secondbest{0.523} \\
& Ours
& 22.52 & \secondbest{0.722} & \best{0.218} & \best{0.106} & \best{0.579} & \best{0.422} & \best{66.79} & \secondbest{5.32} & \best{0.583} \\

\midrule
\multirow{5}{*}{\centering Group3}
& VRT~\cite{liang2024vrt}
& 11.74 & 0.392 & 0.621 & 0.272 & 0.424 & 0.209 & 36.89 & 16.45 & 0.201 \\
& X-Restormer~\cite{chen2024comparative}
& 11.85 & 0.348 & 0.613 & 0.276 & 0.447 & 0.200 & 38.67 & 14.18 & 0.194 \\
& Random Choice
& 13.94 & 0.501 & 0.462 & 0.233 & 0.458 & 0.251 & 43.77 & 6.28 & 0.311 \\
& Qwen3-MoE~\cite{bai2025qwen3vl}
& \secondbest{15.02} & \secondbest{0.536} & \secondbest{0.417} & \secondbest{0.208} & \secondbest{0.481} & \secondbest{0.278} & \secondbest{46.35} & \secondbest{4.97} & \secondbest{0.372} \\
& Ours
& \best{16.03} & \best{0.571} & \best{0.374} & \best{0.191} & \best{0.497} & \best{0.299} & \best{48.99} & \best{3.05} & \best{0.424} \\

\bottomrule
\end{tabular}}
\label{degradation}
\vspace{-10pt}
\end{table*}

\begin{figure*}[t!]
	\centering
    \includegraphics[width=1.0\linewidth]{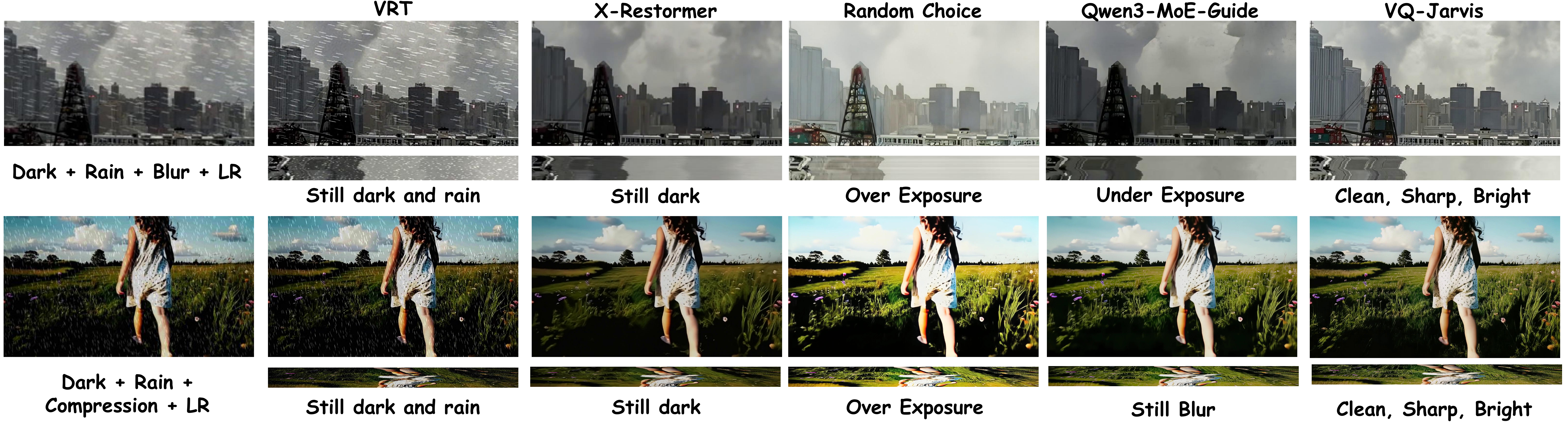}
	\caption{Qualitative comparison between our VQ-Jarvis and other methods. The time profile of each video are shown below.}
	\label{zhuguan}
    \vspace{-10pt}
\end{figure*}

\subsection{Hierarchical Restoration Operator Scheduling}

To efficiently search the best restoration solution, VQ-Jarvis uses a hierarchical operator scheduling strategy including a one-step retrieval and a step-wise greedy search.

\textbf{Prior-guided One-Step Retrieval.} Considering that videos with similar degradations tend to share similar optimal restoration strategies, we attempt to allow the agent to effectively reuse prior restoration memory. The RAG library is constructed offline by applying step-wise comparison using the trained $\mathcal{H}_{\text{judge}}$ to decide optimal restoration trajectories for 1,000 reference videos in the training set. Each entry in the library contains a degraded video and its corresponding best restoration trajectory. For retrieval, the most critical challenge lies in defining a metric space that can effectively capture video quality similarity. Standard CLIP models are primarily designed to measure semantic similarity and are often insensitive to low-level visual artifacts and fine-grained details. To address this limitation, inspired by the RALI framework~\cite{zhao2025reasoning}, we align CLIP embeddings using quality reasoning generated by the Q-Insight, resulting in a quality-aligned embedding space. Given a degraded video, we first sample $N$ frames and compute its average quality-aware image embeddings, namely: 
{\setlength\abovedisplayskip{0.1cm}
\setlength\belowdisplayskip{0.1cm}
\begin{equation}
\mathbf{e}_{\text{emb}}(\mathbf{V}) = \frac{1}{N} \sum_{i=1}^{N} \phi_{\text{emb}}(\mathbf{f}_i),
\end{equation}}where $\{f_i\}_{i=1}^{N}$ are uniformly sampled frames from $\mathbf{V}$, and $\phi_{\text{emb}}(\cdot)$ denotes a quality-aligned embedding model. We then retrieve the Top-3 videos that are most similar to the input video in this metric space and extract their corresponding restoration trajectories. Based on the degradation types predicted by $\mathcal{H}_{\text{deg}}$, we subsequently select the restoration trajectory that best matches the identified degradations. Specifically, given a RAG library $\mathbb{D} = \{(\mathbf{V}_j, \mathcal{T}_j)\}_{j=1}^{M}$, the restoration trajectory is selected as.
{\setlength\abovedisplayskip{0.1cm}
\setlength\belowdisplayskip{0.1cm}
\begin{equation}
\mathcal{T}^* =
\arg\max_{\mathcal{T}_j \in \mathcal{N}_K(\mathbf{V}_{\text{deg}})}
\text{match}\!\left(\mathcal{H}_{\text{deg}}(\mathbf{V}_{\text{deg}}), \mathcal{H}_{\text{deg}}(\mathbf{V}_j)\right), \notag
\end{equation}
\begin{equation}
\mathcal{N}_K(\mathbf{V}_{\text{deg}}) = \arg\max_{\mathbf{V}_j \in \mathcal{D}}^{K}
\text{sim}\!\left(\mathbf{e}_{\text{emb}}(\mathbf{V}_{\text{deg}}), \mathbf{e}_{\text{emb}}(\mathbf{V}_j)\right),
\end{equation}}where $\text{sim}(\cdot,\cdot)$ denotes cosine similarity.

\textbf{Posterior-guided Step-Wise Greedy Search.} For difficult restoration cases, VQ-Jarvis performs a step-wise greedy search. Based on empirical observations and prior knowledge, we adopt a canonical ordering of performing sub-tasks based on extensive experiments and analysis (described in \textbf{Appendix}~\ref{order}), inspired by~\cite{jiang2025multi,zhu2024intelligent}. The overall restoration process is decomposed into a sequence of sub-tasks. At each sub-task, all candidate restoration operators are applied, and their outputs are evaluated by the comparison model using a tournament-style elimination. The optimal operator is selected and applied to produce an intermediate result, which is then passed to the next sub-task. In cases where multiple operators are ranked equally, the operator with the shortest runtime is selected. This greedy procedure continues until all sub-tasks are completed, yielding the final restored video $\mathbf{V}_{\text{recon}}$.

\section{Experimental Results}

\subsection{Implementation Details}
$\mathcal{H}_{\text{deg}}$ and $\mathcal{H}_{\text{judge}}$ are finetuned from Qwen3-VL-8B-Instruct~\cite{bai2025qwen3vl} via GRPO. The degradation perception model is trained in a multi-task manner on the LSVQ training set together with 3,000 synthetically degraded videos from VSR-Compare. The multi-operator judge model is trained on 20K comparison pairs. Both models are trained for approximately 2 days on 8 NVIDIA A100 GPUs. For the RAG embedding, we adopt \textsc{clip-vit-large-patch14-336} as the base encoder and align it using reasoning texts from Q-Insight. We use LlamaIndex for vector retrieval. The pre-defined score threshold $\tau$ is 2.6.

\subsection{Real-World Video Super-Resolution}
To evaluate the performance of our VQ-Jarvis on real-world high-order degradation super-resolution, we evaluate our results on two widely used benchmarks, UDM10~\cite{tao2017detail} and YouHQ40~\cite{zhou2024upscale}. We compare our method against representative state-of-the-art approaches, including SeedVR2~\cite{wang2025seedvr2}, MGLD-VSR~\cite{yang2024motion}, DOVE~\cite{chen2025dove}, Flash-VSR~\cite{zhuang2025flashvsr}, and DOVER-guide~\cite{wu2023exploring}. Here, DOVER-guide denotes a metric-guided baseline that selects the best enhanced result among multiple candidates according to the DOVER score. As shown in Tab.~\ref{realsr}, VQ-Jarvis consistently achieves top or near-top performance across both datasets, especially on non-reference metrics such as CLIPIQA, DOVER, and the overall VQ-Insight score~\cite{zhang2025vq}. Notably, our method outperforms DOVER-guide, indicating that directly optimizing for holistic quality understanding is more effective than metric-based selection alone. These results show that VQ-Jarvis can effectively perform decision-making and restore videos in real-world super-resolution task.
\begin{table}[t]
\centering
\caption{Degradation detection accuracy (\%) comparison between our VQ-Jarvis and other methods.}
\renewcommand{\arraystretch}{1.15}
\resizebox{\linewidth}{!}{
\begin{tabular}{lcccc}
\toprule
Degradation & Qwen3-VL-8B & Qwen3-MoE & Gemini3 Pro & Ours \\
\midrule
Dark           
& 56.07 
& 54.20 
& \secondbest{58.34} 
& \best{83.83} \\

Rain           
& \secondbest{92.31} 
& 89.71 
& 91.02 
& \best{96.40} \\

Blur           
& 67.61 
& 63.24 
& \secondbest{69.15} 
& \best{94.61} \\

Compression    
& 58.30 
& 57.77 
& \secondbest{60.48} 
& \best{87.47} \\

Noise          
& \secondbest{75.51} 
& 66.18 
& 71.62 
& \best{91.35} \\

Low-Resolution 
& 41.30 
& 53.36 
& \secondbest{56.74} 
& \best{94.83} \\

Low-Frame      
& \secondbest{92.13} 
& 84.45 
& 87.96 
& \best{92.27} \\

\midrule
Average        
& 69.03 
& 66.99 
& \secondbest{70.47} 
& \best{91.53} \\
\bottomrule
\end{tabular}}
\label{recognition}
\vspace{-5pt}
\end{table} 
\subsection{Multi-Degradation Video Restoration}

To evaluate the effect of multi-degradation video restoration, we collect 100 source videos including Face, AIGC, REDS~\cite{nah2019ntire}, RealVSR~\cite{yang2021real} testsets, and add 3 representative degradation groups to them via our data pipeline. Here, Group1 includes ``dark + rain + bnc + low resolution'', where ``bnc'' denotes randomly applying one of blur, noise, or compression. Group2 is ``rain + low resolution'', while Group3 contains ``dark + bnc + low-frame''. Tab.~\ref{degradation} reports the average metrics on our constructed benchmark. Our comparative methods include all-in-one video restoration models: VRT~\cite{liang2024vrt} and X-Restormer~\cite{chen2024comparative}, two routing-based baselines: Random Choice, which randomly selects a restoration operator for each sub-task, and Qwen3-MoE, which uses the Qwen3-MoE model as a router.

As reported, our method consistently achieves the best metrics across all three groups. In particular, it shows clear advantages on perceptual quality and holistic evaluation metrics such as CLIPIQA and DOVER, while also maintaining strong distortion-based metrics (LPIPS, DISTS, and Ewarp). Compared with Random Choice and Qwen3-MoE, our approach shows more reliable and robust operator selection, leading to consistently higher-quality restoration under diverse and complex degradation combinations. Fig.~\ref{zhuguan} further shows that our method can accurately identify all degradations and enhance video quality in an efficient manner.

\subsection{Degradation Perception Evaluation}
To evaluate the degradation perception capability of our VQ-Jarvis, we conduct experiments on the degradation recognition tasks. We evaluate on 500 videos with varying degradation levels sampled from our newly constructed VSR-Compare testing set, and compare against strong general MLLMs, including Qwen3-8B-VL, Qwen3-MoE, and Gemini3 Pro. As shown in Tab.~\ref{recognition}, VQ-Jarvis significantly outperforms all competing models in detecting degradations such as dark, rain, and blur, showing great perception capability across all types. 
\begin{table}[t]
\centering
\caption{Ablation study on the key components of our VQ-Jarvis.}
\resizebox{1.\linewidth}{!}{
\renewcommand{\arraystretch}{1.15}
\begin{tabular}{lccccc}
\toprule
Method 
& DISTS$\downarrow$ 
& MANIQA$\uparrow$ 
& Ewarp$\downarrow$ 
& DOVER$\uparrow$ 
& Time~(s)$\downarrow$ \\
\midrule
CLIP-RAG 
& 0.192
& 0.553 
& 3.02 
& 0.396 
& 12.44 \\

Ours-RAG 
& 0.166 
& 0.589 
& 2.88 
& 0.426 
& \best{12.44} \\

Ours-Greedy 
& \best{0.157} 
& \best{0.608} 
& \best{2.72} 
& \secondbest{0.442} 
& 42.65 \\

Ours 
& \secondbest{0.163} 
& \secondbest{0.603} 
& \secondbest{2.80} 
& \best{0.454} 
& \secondbest{17.88} \\
\bottomrule
\end{tabular}}
\label{ablation}
\end{table}

\subsection{Scoring Results of our VQ-Jarvis}

\textbf{Nature Video Scoring.} Furthermore, we conduct video quality scoring experiments on LSVQ-Test~\cite{ying2021patch}, LSVQ-1080p~\cite{ying2021patch}, LIVE-VQC~\cite{sinno2018large}, and KoNViD-1k~\cite{hosu2017konstanz}. As reported in Tab.~\ref{tab:performance_comparison}, VQ-Jarvis consistently surpasses traditional VQA methods such as DOVER~\cite{wu2023exploring} and Modular-VQA~\cite{wen2024modular}, as well as reasoning-based MLLMs like VQ-Insight~\cite{zhang2025vq}. These results indicate that VQ-Jarvis can more accurately perceive degradation characteristics in input videos and make more reliable quality judgments, highlighting its strong generalization and fine-grained degradation reasoning ability. The predicted score is used as a criterion to choose the one-step retrieval mode or the step-wise greedy search mode. 

\textbf{Fine-Grained UGC Video Scoring.} By further training the degradation-aware model $\mathcal{H}_{\text{deg}}$ with the multi-dimensional scoring rewards proposed in VQ-Insight~\cite{zhang2025vq}, our method can achieve robust multi-dimensional quality assessment capability on UGC videos~\cite{duan2025finevq}. Tab.~\ref{tab:performance_comparison} reports the quantitative comparison across six quality dimensions, including Color, Noise, Artifact, Blur, Temporal, and Overall. We compare our method with a wide range of representative video quality assessment approaches, covering traditional handcrafted metrics (NIQE), learning-based image/video quality models (VIDEVAL, RAPIQUE), as well as recent deep video quality models such as VSFA, GSTVQA, SimpleVQA, FAST-VQA, and DOVER.

Overall, VQ-Jarvis consistently achieves the best performance across almost all dimensions, significantly outperforming prior methods in terms of both PLCC and SRCC. Compared with the strongest baseline FineVQ, VQ-Jarvis shows clear improvements on Color, Artifact, Blur, and Overall dimensions, indicating stronger capability in capturing complex perceptual degradations and holistic video quality. In particular, the gains on Artifact and Blur demonstrate that VQ-Jarvis can better handle multi-degradation scenarios where structural distortions and restoration artifacts co-exist. While FineVQ exhibits competitive performance on Noise and Temporal dimensions, VQ-Jarvis maintains comparable temporal consistency while achieving superior overall correlation with human preferences. These results validate that integrating accurate degradation recognition with coordinated sub-task execution enables VQ-Jarvis to produce more natural, clean, and perceptually consistent video quality assessments than FineVQ and other competing approaches.

\begin{table*}[h]
    \centering
    \renewcommand{\arraystretch}{1.3}
    \caption{PLCC / SRCC comparisons on natural video quality assessment benchmarks.}
    \resizebox{0.9\linewidth}{!}{
    \begin{tabular}{c|cccc}
        \shline
        Model &
        \begin{tabular}[c]{@{}c@{}}LSVQ-Test\\(PLCC / SRCC)\end{tabular} &
        \begin{tabular}[c]{@{}c@{}}LSVQ-1080p\\(PLCC / SRCC)\end{tabular} &
        \begin{tabular}[c]{@{}c@{}}LIVE-VQC\\(PLCC / SRCC)\end{tabular} &
        \begin{tabular}[c]{@{}c@{}}KonViD-1k\\(PLCC / SRCC)\end{tabular} \\
        \hline

        Fast-VQA~\cite{wu2022fast} &
        0.878 / 0.874 &
        0.810 / 0.765 &
        0.815 / 0.769 &
        0.857 / 0.859 \\

        Minimalist-VQA~\cite{sun2024analysis} &
        0.872 / 0.880 &
        0.818 / 0.769 &
        0.812 / 0.765 &
        0.861 / 0.859 \\

        DOVER~\cite{wu2023exploring} &
        0.886 / 0.888 &
        0.828 / 0.787 &
        0.819 / 0.771 &
        \secondbest{0.883} / \secondbest{0.890} \\

        Modular-VQA~\cite{wen2024modular} &
        \secondbest{0.891} / \best{0.894} &
        \best{0.844} / \secondbest{0.791} &
        0.825 / 0.783 &
        \secondbest{0.884} / 0.878 \\

        Q-Align~\cite{wu2024q} &
        0.882 / 0.883 &
        0.833 / 0.758 &
        0.813 / 0.777 &
        0.876 / 0.865 \\

        Q-Instruct~\cite{wu2024qinstruct} &
        0.580 / 0.602 &
        0.640 / 0.644 &
        0.673 / 0.660 &
        0.520 / 0.492 \\

        VQA$^2$~\cite{jia2024vqa} &
        0.856 / 0.882 &
        0.819 / 0.760 &
        0.823 / 0.776 &
        0.844 / 0.833 \\

        VQ-Insight~\cite{zhang2025vq} &
        0.876 / 0.875 &
        0.823 / 0.786 &
        \secondbest{0.835} / \secondbest{0.790} &
        \secondbest{0.884} / 0.875 \\

        \textbf{VQ-Jarvis (Ours)} &
        \best{0.893} / \secondbest{0.893} &
        \secondbest{0.840} / \best{0.809} &
        \best{0.862} / \best{0.820} &
        \best{0.899} / \best{0.891} \\

        \shline
    \end{tabular}}
    \label{tab:performance_comparison}
\end{table*}

\begin{table*}[t]
\centering
\caption{Performance comparison of different quality dimensions on the FineVQ dataset. PLCC and SRCC are reported. Best and second-best results are highlighted.}
\renewcommand{\arraystretch}{1.3}
\resizebox{\textwidth}{!}{
\begin{tabular}{lcccccccccccc}
\toprule
\multirow{2}{*}{Method} 
& \multicolumn{2}{c}{Color} 
& \multicolumn{2}{c}{Noise} 
& \multicolumn{2}{c}{Artifact} 
& \multicolumn{2}{c}{Blur} 
& \multicolumn{2}{c}{Temporal} 
& \multicolumn{2}{c}{Overall} \\
& PLCC & SRCC & PLCC & SRCC & PLCC & SRCC & PLCC & SRCC & PLCC & SRCC & PLCC & SRCC \\
\midrule
NIQE~\cite{mittal2012making} 
& 0.2368 & 0.3273 & 0.1417 & 0.2682 & 0.1649 & 0.3006 & 0.1870 & 0.3236 & 0.1630 & 0.2777 & 0.2019 & 0.3192 \\
VIDEVAL~\cite{tu2021ugc} 
& 0.6943 & 0.6922 & 0.6514 & 0.6912 & 0.7370 & 0.7440 & 0.7637 & 0.7610 & 0.7123 & 0.7174 & 0.7307 & 0.7310 \\
RAPIQUE~\cite{tu2021rapique} 
& 0.6330 & 0.6203 & 0.5607 & 0.6048 & 0.6276 & 0.6258 & 0.6717 & 0.6572 & 0.5429 & 0.5554 & 0.6501 & 0.6379 \\
VSFA~\cite{li2019quality} 
& 0.7837 & 0.7617 & 0.7278 & 0.7635 & 0.8170 & 0.8006 & 0.8001 & 0.7772 & 0.7018 & 0.7276 & 0.7929 & 0.7730 \\
GSTVQA~\cite{chen2021learning} 
& 0.7761 & 0.7747 & 0.7448 & 0.7883 & 0.8187 & 0.8121 & 0.8202 & 0.8101 & 0.7093 & 0.7533 & 0.7825 & 0.7834 \\
SimpleVQA~\cite{sun2022deep} 
& 0.8058 & 0.8086 & 0.7634 & 0.8070 & 0.8487 & 0.8465 & 0.8519 & 0.8466 & 0.7417 & 0.7746 & 0.8358 & 0.8311 \\
FAST-VQA~\cite{wu2022fast} 
& 0.8183 & 0.8017 & 0.7758 & 0.8093 & 0.8328 & 0.8176 & 0.8513 & 0.8352 & 0.7393 & 0.7560 & 0.8474 & 0.8348 \\
DOVER~\cite{wu2023exploring} 
& 0.8311 & 0.8244 & 0.7424 & 0.8018 & 0.8289 & 0.8265 & 0.8355 & 0.8404 & 0.7569 & 0.7664 & 0.8393 & 0.8422 \\
FineVQ~\cite{duan2025finevq} 
& \secondbest{0.8527} & \secondbest{0.8495}
& \secondbest{0.7986} & \best{0.8444}
& \secondbest{0.8921} & \secondbest{0.8852}
& \secondbest{0.8833} & \secondbest{0.8711}
& \secondbest{0.7597} & \best{0.8085}
& \secondbest{0.8891} & \secondbest{0.8834} \\ \midrule

\textbf{VQ-Jarvis (Ours)} 
& \best{0.8937} & \best{0.8800}
& \best{0.8055} & \secondbest{0.8271}
& \best{0.8991} & \best{0.8927}
& \best{0.9166} & \best{0.9100}
& \best{0.7814} & \secondbest{0.7952}
& \best{0.9081} & \best{0.9007} \\
\bottomrule
\end{tabular}}
\label{tab:fineq_vqjarvis}
\end{table*}

\subsection{Ablation Studies}
To validate the effect of each component, we conduct experiments on the RealVSR subset of the VSR-Compare testing set, using the 3 degradation groups. Tab.~\ref{ablation} reports the results of different decision strategies. The testing time is average running time of reconstructing a degraded video. Compared with CLIP-RAG, which mainly focuses on high-level semantic information, Ours-RAG uses visual quality–aligned embeddings to more accurately retrieve videos with similar degradations, thus enabling more precise operator selection. Ours-RAG, relying solely on retrieval-based search, achieves faster inference but exhibits a noticeable drop in reconstruction quality, indicating that RAG alone is insufficient for fine-grained restoration decisions. In contrast, Ours-Greedy, which performs exhaustive greedy search over candidate operators, attains the best quality scores but incurs a significant computational overhead, with over 40 seconds per video. By integrating predicted quality scores to dynamically balance retrieval and greedy exploration, Ours effectively achieves competitive restoration quality and much lower inference time. It shows that our hybrid design provides a favorable trade-off between restoration quality and efficiency.

\subsection{Time Complexity Analysis}
We analyze the computational complexity of the proposed framework under two inference strategies: one-step retrieval and step-wise greedy search. 
Let $K$ denote the number of sub-tasks in the restoration pipeline, and $N$ the number of candidate restoration operators per sub-task. $T_{\text{cmp}}$ the cost of a single pairwise comparison, $T_{\text{rag}}$ the retrieval cost of the RAG module, and $T_{\text{det}}$ the shared cost of degradation perception.
Let $T_{k,i}$ denote the execution time of the $i$-th restoration operator for the $k$-th sub-task (e.g., deraining, low-light enhancement, super-resolution), where $k \in \{1,\dots,K\}$ and $i \in \{1,\dots,N\}$.
Accordingly, the average operator execution time is defined as $T_{\text{op}} = \frac{1}{K} \sum_{k=1}^{K} \mathbb{E}_{i}[T_{k,i}]$,
while the worst-case execution time for a sub-task is given by $T_{\max} = \max_{k,i} T_{k,i}$.

For the RAG-based pipeline, the framework performs a single retrieval followed by the execution of one operator per sub-task, leading to a time complexity of
$\mathcal{O}\!\left(T_{\text{rag}} + K\,T_{\text{op}}\right)$.
In contrast, the greedy search strategy requires $(N-1)$ pairwise comparisons for each sub-task and executes the most time-consuming operator (since the restoration operators within each sub-task are executed in parallel), resulting in a complexity of
$\mathcal{O}\!\left(K\left(N\,T_{\text{cmp}} + T_{\max}\right)\right)$.
Combining both strategies with the shared perception cost, the expected total runtime of the framework can be expressed as $\mathcal{O}~\!\big(
T_{\text{det}}
+ \rho\left(T_{\text{rag}} + K\,T_{\text{op}}\right)
+ (1-\rho)\,K\left(N\,T_{\text{cmp}} + T_{\max}\right)
\big)$, 
where $\rho \in [0,1]$ denotes the probability of triggering the RAG-based pipeline. Therefore, when $\rho$ is large, the overall runtime of the agent is dominated by the RAG-based execution path and can be approximated as
$
\mathcal{O}\!\left(
T_{\text{det}} + T_{\text{rag}} + \sum_{k=1}^{K} T_{k,i_k}
\right),$
where $i_k$ denotes the operator selected for the $k$-th sub-task.
In expectation, this further simplifies to
$\mathcal{O}\!\left(
T_{\text{det}} + T_{\text{rag}} + K\,T_{\text{op}}
\right)$,
indicating a near-linear dependence on the number of sub-tasks. The detailed runtime of each key module is listed in Tab.~\ref{tab:runtime_analysis}.

\begin{table*}[h]
\centering
\caption{Runtime breakdown of different components and restoration operators in VQ-Jarvis.}
\label{tab:runtime_analysis}
\renewcommand{\arraystretch}{1.15}
\resizebox{0.8\linewidth}{!}{
\begin{tabular}{lccc}
\toprule
\textbf{Module / Operator} & \textbf{Task} & \textbf{Symbol} & \textbf{Time (s)} \\
\midrule
\multicolumn{4}{l}{\textit{Decision and Perception Overhead}} \\
\midrule
Degradation Perception Model & All & $T_{\text{det}}$ & 2.45 \\
Multiple Operator Judge Model & All & $T_{\text{cmp}}$ & 2.81 \\
RAG Retrieval & All & $T_{\text{rag}}$ & 0.35 \\
\midrule
\multicolumn{4}{l}{\textit{Deraining Operators (30$\times$180p videos)}} \\
\midrule
RainMamba & Deraining & $T_{1, 1}$ & 0.56 \\
3D-UNet-R & Deraining & $T_{1, 2}$ & 4.24 \\
\midrule
\multicolumn{4}{l}{\textit{Low-Light Enhancement Operators (30$\times$180p videos)}} \\
\midrule
Adaptive Gamma Correction & Low-Light Enhancement & $T_{2,1}$ & 0.85 \\
InstructIR & Low-Light Enhancement & $T_{2,2}$ & 1.13 \\
3D-UNet-L & Low-Light Enhancement & $T_{2,3}$ & 4.24 \\
\midrule
\multicolumn{4}{l}{\textit{Restoration Operator (30$\times$720p videos)}} \\
\midrule
FlashVSR & SR / Deblur / Decompression / Denoise & $T_{3, 1}$ & 3.28 \\
DOVE & SR / Deblur / Decompression / Denoise & $T_{3,2}$ & 10.42 \\
SeedVR-3B & SR / Deblur / Decompression / Denoise & $T_{3, 3}$ & 15.75 \\
RealBasicVSR++ & Super-Resolution~(SR) & $T_{3, 4}$ & 0.94 \\ \midrule
\multicolumn{4}{l}{\textit{Frame Interpolation Operator (10$\times$720p videos)}} \\
GIMM-VFI & Frame Interpolation & $T_{4, 1}$ & 5.23 \\
\bottomrule
\end{tabular}}
\end{table*}


\subsection{Sensitivity Analysis}
Tab.~\ref{tab:threshold_sensitivity} reports the sensitivity of VQ-Jarvis with respect to the threshold score $\tau$. 
Overall, the performance varies smoothly as $\tau$ changes, indicating that our method is not overly sensitive to the specific choice of the threshold.
Across a wide range of values ($\tau \in [2.4, 3.6]$), VQ-Jarvis maintains competitive performance on all quality metrics, with no abrupt degradation observed. Specifically, smaller thresholds (e.g., $\tau=2.4$) bias the system toward retrieval-based restoration, leading to faster inference but slightly inferior perceptual quality.
Larger thresholds (e.g., $\tau=3.6$) favor more aggressive multi-step greedy search, yielding marginal gains on certain metrics at the cost of substantially increased runtime.
The default setting $\tau=2.6$ achieves a favorable balance between restoration quality and efficiency, delivering near-optimal performance across all metrics while keeping the inference time moderate.
These results demonstrate the robustness of VQ-Jarvis to the threshold selection and justify our choice of $\tau=2.6$ as the default configuration.

\begin{table*}[h]
\centering
\caption{Sensitivity analysis of the pre-defined threshold score $\tau$ in VQ-Jarvis.}
\resizebox{0.7\linewidth}{!}{
\renewcommand{\arraystretch}{1.15}
\begin{tabular}{lccccc}
\toprule
Threshold $\tau$
& DISTS$\downarrow$
& MANIQA$\uparrow$
& Ewarp$\downarrow$
& DOVER$\uparrow$
& Time~(s)$\downarrow$ \\
\midrule

$\tau = 2.4$
& 0.165
& 0.592
& 2.86
& 0.430
& \best{13.95} \\

$\tau = 2.6$ (Ours)
& 0.163
& \secondbest{0.603}
& \secondbest{2.80}
& \best{0.454}
& \secondbest{17.88} \\

$\tau = 3.0$
& \secondbest{0.160}
& 0.600
& 2.78
& \secondbest{0.446}
& 29.04 \\

$\tau = 3.6$
& \best{0.159}
& \best{0.607}
& \best{2.78}
& 0.444
& 39.93 \\

\bottomrule
\end{tabular}}
\label{tab:threshold_sensitivity}
\end{table*}



\subsection{Reconstruction Order Experience}
\label{order}
Following prior studies on reconstruction order determination, most existing works—such as AgenticIR~\cite{zhu2024intelligent}, MAIR~\cite{jiang2025multi}, and 4KAgent~\cite{zuo20254kagent}—primarily rely on feeding a large language model with experience accumulated from a series of reconstruction cases to perform task scheduling.
Our approach follows a similar philosophy. However, due to the introduction of a set of composite operators (SeedVR2~\cite{wang2025seedvr2}, FlashVSR~\cite{zhuang2025flashvsr}), tasks such as deblurring, decompression, denoising, and super-resolution can often be accomplished in a single step by one operator. As a result, the reconstruction order becomes significantly simpler to determine.

Specifically, we denote random combinations of blur, noise, and compression as BNC, super-resolution as SR, and frame interpolation as FI.
As shown in Tab.~\ref{tab:degradation_order}, we conduct experiments on 50 videos with multiple degradations, considering two degradation settings:
(1) rain + dark + BNC + low resolution, and
(2) dark + BNC + frame dropping.
For consistency, we fix the tools used for each task as follows: 3D-UNet-L for restoration, Gamma correction for low-light enhancement, DOVE for deraining, and GIMM-VFI for frame interpolation.

First, we investigate the ordering between scene degradations (dark, rain) and compression-related degradations (BNC+SR).
Since operators such as SeedVR2 and FlashVSR (representative BNC+SR methods) leverage strong diffusion priors and generally produce high-quality outputs, performing deraining and low-light enhancement before super-resolution leads to significantly better visual quality.
This observation is validated by comparing cases (a) and (c) in Tab.~\ref{tab:degradation_order}. Therefore, BNC/SR operators should be applied after dark and rain restoration.

Next, we explore the ordering between deraining and low-light enhancement.
Video deraining is a particularly challenging task, and applying low-light enhancement first may introduce artifacts that interfere with the deraining network’s ability to identify rain streaks, resulting in incomplete rain removal.
This effect is confirmed by comparing cases (a) and (b). Hence, the low-light enhancement operator should be placed after deraining.

\begin{table*}[t]
\centering
\caption{Quantitative comparison under different degradation orderings.}
\resizebox{1\linewidth}{!}{
\renewcommand{\arraystretch}{1.3}
\label{tab:degradation_order}
\begin{tabular}{clccccccc}
\shline
Case & Restoration Order
& LPIPS~$\downarrow$ 
& DISTS~$\downarrow$ 
& MANIQA~$\uparrow$ 
& CLIPIQA~$\uparrow$ 
& MUSIQ~$\uparrow$ 
& Ewarp~$\downarrow$ 
& DOVER~$\uparrow$ \\
\hline
(a) & Rain + Dark + (BNC + SR) 
& \best{0.245} & \best{0.141} & \secondbest{0.461} & \best{0.392} & \best{56.025} & \best{1.51} & \best{0.571} \\

(b) & Dark + Rain + (BNC + SR) 
& \secondbest{0.264} & \secondbest{0.144} & 0.422 & \secondbest{0.360} & 50.345 & 2.14 & \secondbest{0.568} \\

(c) & (BNC + SR) + Rain + Dark 
& 0.332 & 0.178 & \best{0.468} & 0.310 & \secondbest{52.210} & \secondbest{1.93} & 0.452 \\

\hline
(d) & Dark + BNC + FI 
& \best{0.349} & \best{0.166} & 0.455 & 0.305 & 49.557 & \best{1.04} & \best{0.459} \\

(e) & FI + Dark + BNC 
& 0.325 & \secondbest{0.168} & \secondbest{0.473} & \secondbest{0.331} & \secondbest{50.537} & 1.65 & 0.442 \\

(f) & Dark + FI + BNC 
& \secondbest{0.332} & 0.171 & \best{0.475} & \best{0.345} & \best{51.339} & \secondbest{1.31} & \secondbest{0.445} \\
\shline
\end{tabular}}
\end{table*}

We then examine the relationship between FI and BNC/SR.
Comparing cases (d) and (e), we observe that although applying BNC at the final stage improves spatial quality (e.g., CLIPIQA and MUSIQ), it degrades temporal consistency (e.g., Ewarp and DOVER).
More importantly, BNC operators are typically computationally expensive diffusion-based models, and running them on high-frame-rate intermediate results after frame interpolation dramatically increases the overall pipeline runtime.
Therefore, frame interpolation should be performed after BNC operators.

Finally, we analyze the ordering between FI and dark/rain restoration.
Comparing cases (e) and (f), we find that performing low-light enhancement before or after frame interpolation results in comparable spatial quality, while applying frame interpolation at a later stage yields better temporal performance.
A similar trend is observed for the relationship between FI and deraining.
Consequently, frame interpolation should be applied after dark and rain restoration.

In summary, we derive the following reconstruction order experience:

\textbf{(1) Deraining should precede low-light enhancement}, as early enhancement introduces artifacts that hinder rain removal;

\textbf{(2) BNC/SR operators should be applied after dark and rain restoration}, since their strong diffusion priors benefit from cleaner inputs; 

\textbf{(3) Frame interpolation should be performed last, after both scene restoration and BNC/SR}, to achieve better temporal consistency and avoid excessive computational overhead.

It is important to note that the above reconstruction experience are specific to the operators and experimental setting considered in this work and should not be interpreted as universally optimal rules.
In fact, our framework can naturally incorporate a data-driven scheduling strategy similar to AgenticIR, where reconstruction experiences or statistical summaries from multiple cases are provided to a large language model for dynamic decision-making and task scheduling.
This extension is fully compatible with our approach and does not contradict our design. The core innovation of our method lies not in replacing such scheduling strategies, but in enhancing the agent’s ability to perceive subtle differences among reconstruction outcomes and to efficiently select the most appropriate operator for each sub-task, enabling more precise and cost-effective restoration decisions.

\section{Conclusion}

In this paper, we present VQ-Jarvis, a retrieval-augmented, all-in-one intelligent video restoration agent designed for complex real-world degradations. By unifying quality-aware perception, preference-based comparison, and adaptive operator scheduling, VQ-Jarvis equips the agent with sharp vision to accurately perceive degradations and subtle quality differences, as well as fast thought to efficiently identify effective restoration trajectories, achieving strong restoration performance with high computational efficiency. Another key contribution of our work is VSR-Compare, the first large-scale video paired enhancement dataset, which enables reliable training of degradation perception and multi-operator judging models. We believe that VQ-Jarvis represents an important step toward experience-driven and decision-centric video restoration, and that the principles introduced in this work—quality-aligned perception, preference-based supervision, and retrieval-augmented reasoning—can inspire future research on intelligent and adaptive vision systems beyond restoration.

\newpage
\begin{appendices}
\section{Limitations}

\textbf{First}, VQ-Jarvis relies on several strong diffusion-based restoration operators to achieve high-quality video reconstruction. While these operators provide superior perceptual results, they are inherently computationally expensive, leading to relatively slow inference speed in some cases. Importantly, this limitation originates from the restoration operators themselves rather than the proposed scheduling strategy. In fact, the RAG-based scheduling mechanism introduced in VQ-Jarvis achieves near-linear complexity with respect to the number of degradations, demonstrating high efficiency in operator selection.

\textbf{Second}, the current operator pool, although diverse and effective, is not exhaustive. As video restoration algorithms continue to evolve, the operator pool can be further expanded and refined to include more specialized and efficient restoration methods. We believe that VQ-Jarvis is naturally extensible in this regard, and future improvements in restoration models can be seamlessly integrated to further enhance its performance and flexibility.

\section{More Visualized Results}

\subsection{Multi-Degradation Restoration Results}

\begin{figure*}[t!]
	\centering
    \includegraphics[width=1.0\linewidth]{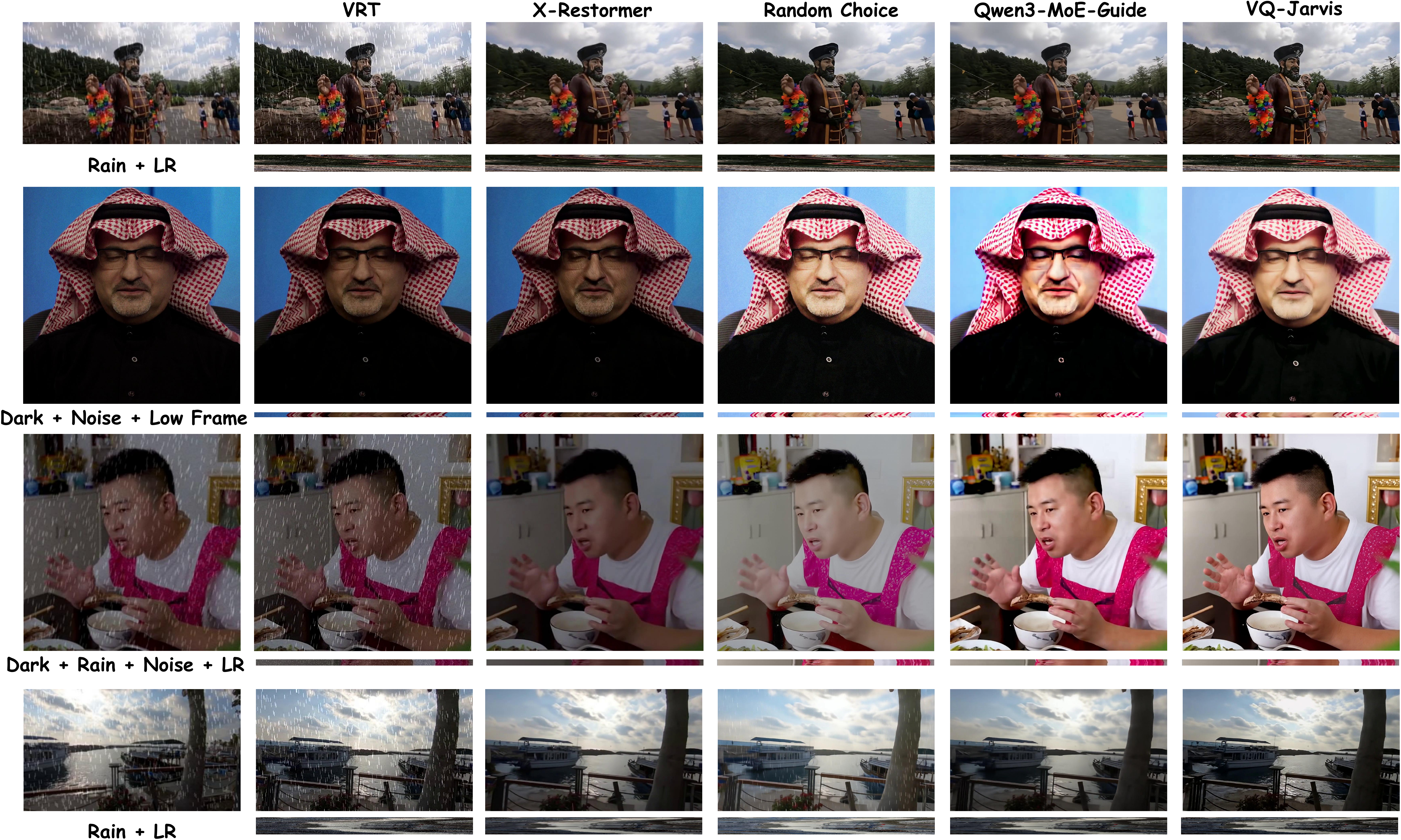}
	\vspace{-10pt}
	\caption{Visualized results of our VQ-Jarvis and other competing methods on multi-degradation video restoration methods.}
	\label{zhuguan3}
    \vspace{-10pt}
\end{figure*}

We present more multi-degradation restoration results in Fig.~\ref{zhuguan3}. Existing methods either fail to fully identify all degradation types or only partially address the restoration tasks, leading to residual rain streaks, blur, noise, or visible artifacts.
In contrast, VQ-Jarvis accurately recognizes all present degradations and executes the corresponding restoration subtasks in a coordinated manner, effectively handling complex combinations such as rain, low resolution, darkness, compression, and noise.
Therefore, VQ-Jarvis produces videos that are visually more natural, clean, and sharp, with well-preserved details and without noticeable blur, artifacts, or temporal inconsistencies.

\subsection{Visualized Results of our Multiple operator Judge and Degradation Perception Model} 

\begin{figure*}[t!]
	\centering
    \includegraphics[width=0.8\linewidth]{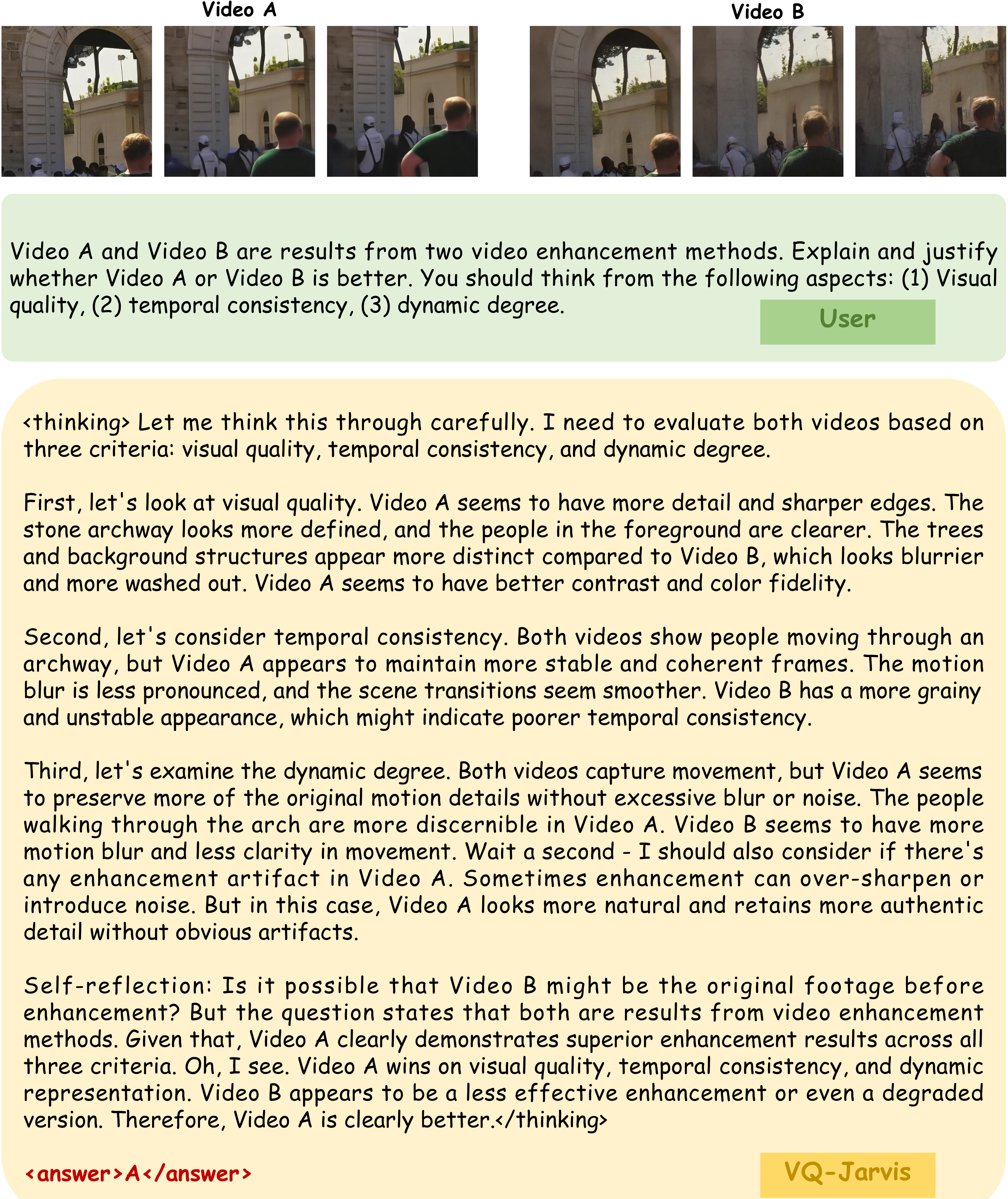}
	\caption{Visualized results of the paired comparison results of our VQ-Jarvis.}
	\label{comp}
    \vspace{-10pt}
\end{figure*}

\begin{figure*}[t!]
	\centering
    \includegraphics[width=0.8\linewidth]{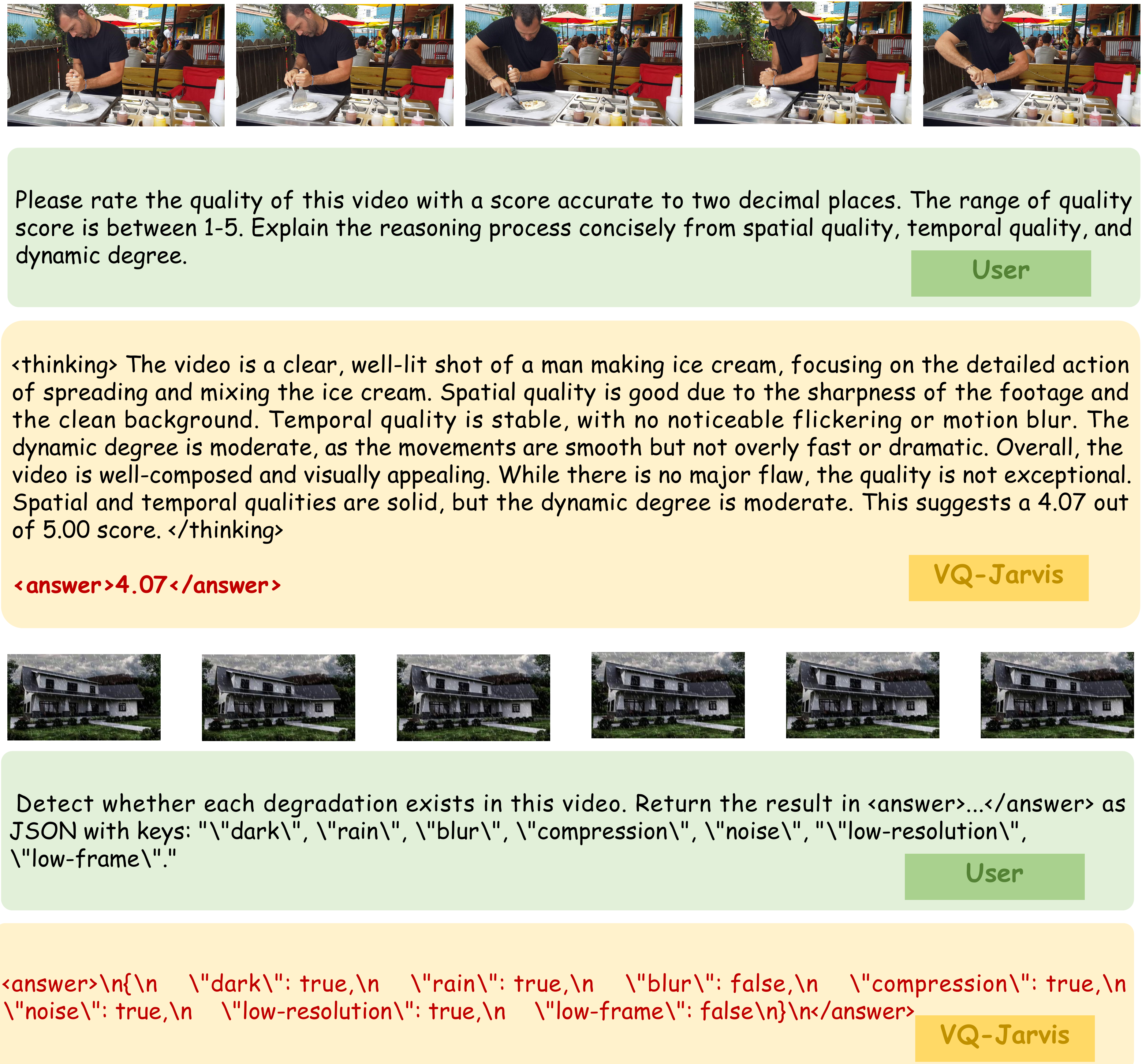}
	\caption{Visualized results of the video quality scoring and degradation detection using our VQ-Jarvis.}
	\label{score}
    \vspace{-10pt}
\end{figure*}

We present the visualized examples of our VQ-Jarvis for video scoring, degradation detection and preference comparison in Fig.~\ref{comp} and Fig.~\ref{score}.

\section{More Details of VSR-Compare}
\subsection{Degradation Operators}
\label{sec:degrade}
We design a diverse set of degradation operators to simulate realistic video quality degradations and maintain temporal consistency. Our degradation pipeline is inspired by AgenticIR~\cite{zhu2024intelligent}, DOVE~\cite{chen2025dove}, and MoA-VR~\cite{liu2025moa}.
For each video, degradation parameters are sampled once at the video level and kept fixed across all frames to ensure temporal consistency and avoid temporal flickering.
The degradation operators and their corresponding severity ranges are defined as follows:

\begin{itemize}
    \item \textbf{Low Resolution.}
    Spatial resolution degradation is simulated by downsampling the input frames by a factor of $4\times$ using bicubic interpolation, optionally followed by resizing back to the original resolution.
    This models resolution loss caused by camera limitations or transmission constraints.

    \item \textbf{Darkness.}
    Low-light conditions are modeled using three strategies: constant intensity shift, gamma correction, and linear mapping.
    For constant shift, the intensity reduction is sampled from $[30, 50]$.
    For gamma correction, the gamma value is sampled from $[0.5, 0.7]$.
    For linear mapping, the target maximum intensity is sampled from $[100, 150]$.

    \item \textbf{Noise.}
    Sensor noise is simulated using either Gaussian noise or Poisson noise.
    For Gaussian noise, the noise standard deviation is sampled from $[20, 50]$.
    For Poisson noise, the noise scale factor is sampled from $[1, 3]$.
    The noise type and magnitude are sampled once per video and applied consistently across all frames.

    \item \textbf{JPEG Compression Artifacts.}
    Compression artifacts are introduced via JPEG encoding with quality factors uniformly sampled from $[10, 30]$, corresponding to severe compression commonly observed in real-world videos.

    \item \textbf{Defocus / Gaussian Blur.}
    Optical blur is modeled using either defocus blur or Gaussian blur.
    For Gaussian blur, the standard deviation $\sigma$ is sampled from severity-dependent ranges, spanning approximately $[0.6, 8.0]$, with kernel sizes adaptively determined by $\sigma$.
    For defocus blur, disk-shaped kernels are constructed with radii ranging from $3$ to $35$ pixels, followed by mild Gaussian smoothing to suppress aliasing.
    These settings cover a broad spectrum of out-of-focus effects encountered in real-world videos.

    \item \textbf{Rain.}
    Rain streaks are synthesized by sampling a sparse random noise map, followed by local smoothing to form rain seeds.
A directionally oriented convolution kernel is then applied to stretch these seeds into elongated rain streaks.
The rain intensity is sampled from $[50, 100]$, while the rain orientation angle is randomly sampled from a fixed range (e.g., $[-30^\circ, 30^\circ]$) and kept constant at the video level to ensure temporal consistency across frames.

    \item \textbf{Video Compression.}
    Temporal compression artifacts are simulated at the video level using standard video codecs (e.g., H.264 or MPEG-4).
    The target bitrate is sampled from $[50\text{k}, 200\text{k}]$, ensuring temporally consistent compression artifacts without frame-wise flickering.

    \item \textbf{Frame Skip.}
    Temporal sparsity is introduced by uniformly skipping frames with a fixed stride (e.g., $4\times$ subsampling), simulating low-frame-rate capture or transmission dropouts.

    \item \textbf{BNC (Blur--Noise--Compression).}
    To model compound degradations frequently observed in real-world videos, we introduce a composite BNC operator.
    One degradation type is randomly selected from defocus blur, noise, JPEG compression, or video compression, with its severity sampled from the corresponding ranges defined above.
\end{itemize}

\subsection{Restoration Operators}
\label{sec:operators}
To construct a diverse and capable operator pool, we incorporate both state-of-the-art restoration models and lightweight in-house networks, covering deraining, low-light enhancement, super-resolution, and general restoration tasks.

\paragraph{Deraining.}    For video deraining, we adopt RainMamba~\cite{wu2024rainmamba}, a recent Mamba-based video deraining model that explicitly models long-range temporal dependencies, together with an in-house trained 3D-UNet-R.

\begin{itemize}
    \item \textbf{RainMamba}: RainMamba~\citep{wu2024rainmamba} is a video deraining model based on State Space Models (SSMs). It adopts the Mamba architecture to process video sequences with linear computational complexity. RainMamba is further fine-tuned on degradations sampled from the VSR-Compare dataset to align it with our degradation setting. RainMamba excels at handling dynamic rain streaks with strong temporal consistency; however, its performance degrades when rain is accompanied by other degradations such as blur and compression, indicating limited robustness under compound degradation settings.

    \item \textbf{3D-UNet-R}: Our 3D-UNet-R follows a fully convolutional 3D U-Net architecture, operating on short video clips and producing restored frames in a single forward pass, offering an efficient alternative for deraining under severe degradations. 3D-UNet-R adopt a deep 3D U-Net architecture that performs spatial downsampling while preserving the temporal dimension.
    The network consists of 6 encoder-decoder stages with a widened bottleneck to enhance representational capacity.
    These models are trained on 1,000 video clips sampled from the DOVE dataset~\cite{chen2025dove}, where degradations are synthetically applied to generate paired training data.
    During inference, a sliding-window strategy is used along the temporal dimension, and large frames are processed using tiled inference with overlapping patches.
    The final output frame is obtained from the center frame of each temporal clip, and overlapping spatial predictions are merged by weighted averaging to ensure seamless reconstruction.
\end{itemize}
    
\paragraph{Low-Light Enhancement.}
    For low-light enhancement, we consider both classical and learning-based approaches.
    \begin{itemize}
    
    \item \textbf{Gamma Correction:} We employ adaptive gamma correction, where the gamma value is adjusted according to the global luminance of the input video. Although fast to compute, gamma correction provides limited enhancement capacity and may lead to under-exposure and minor artifacts.

    \item \textbf{InstructIR:} InstructIR~\citep{conde2024instructir} is an instruction-driven image restoration model that combines a convolutional image backbone with a text encoder to condition restoration behavior on textual prompts.
The encoded instruction is injected into intermediate feature representations via a lightweight conditioning mechanism to enable task-adaptive enhancement.
Although InstructIR enables fast enhancement, the restored videos may exhibit artifacts, temporal inconsistency, and occasional over-exposure under complex degradation settings.
        
        \item \textbf{3D-UNet-L:} In addition, we train an in-house 3D-UNet-L on the sampled 1,000 videos from the DOVE set~\cite{chen2025dove} using the same architecture as 3D-UNet-R, tailored for low-light enhancement. While this operator provides strong overall enhancement performance, tiling artifacts may occasionally appear when handling ultra-high-resolution inputs.

    \end{itemize}

\paragraph{Video Super-Resolution}For super-resolution, we incorporate a set of recent real-world video super-resolution models with complementary characteristics.

\begin{itemize}
    \item \textbf{SeedVR2-3B:} SeedVR2~\citep{wang2025seedvr2} is a one-step diffusion-based video restoration model obtained by adversarial post-training from a pretrained diffusion backbone.
It introduces adaptive window attention and tailored training losses to enable high-resolution video restoration within a single denoising step.
SeedVR2 is particularly effective on cartoon-style videos and tends to generate rich fine-grained details, but may occasionally produce over-sharpened results.

    \item  \textbf{FlashVSR:} FlashVSR~\citep{zhuang2025flashvsr} is a one-step diffusion-based streaming video super-resolution framework designed for real-time inference.
It employs a multi-stage distillation pipeline together with locality-constrained sparse attention and a lightweight conditional decoder to enable efficient high-resolution video restoration.
Benefiting from a strong diffusion backbone, FlashVSR achieves high generation quality while maintaining fast inference speed.

    \item \textbf{DOVE:} DOVE~\citep{chen2025dove} is a one-step diffusion-based video super-resolution model obtained by fine-tuning a pretrained video diffusion backbone (CogVideoX).
It adopts a latent--pixel two-stage training strategy to adapt the diffusion model to real-world video super-resolution.
In practice, the generated video quality may occasionally degrade, and blur artifacts are not always fully removed.

    \item \textbf{MGLD-VSR:} MGLD-VSR~\citep{yang2024motion} is a diffusion-based video super-resolution method that operates in the latent space and incorporates motion-guided constraints to steer the diffusion process across frames.
It optimizes the latent sampling trajectory using motion information and further enhances temporal modeling with additional temporal modules in the decoder.
In practice, MGLD-VSR suffers from relatively slow inference speed and limited temporal consistency, but can produce high-quality reconstructions on videos with mild degradations. Because we observe that MGLD-VSR is extremely slow during inference and often delivers unsatisfactory reconstruction quality, we only include it when constructing VSR-Compare to enrich positive and negative samples and increase data diversity. In contrast, \textbf{MGLD-VSR is not incorporated into the prediction pipeline of VQ-Jarvis}.

    \item \textbf{RealBasicVSR++:} RealBasicVSR++~\citep{chan2022basicvsr++} is a classical reconstruction-based baseline. Compared with diffusion-based approaches, it offers much faster reconstruction and performs well on naturally degraded videos with relatively mild degradation.

\end{itemize}
    
\paragraph{Deblurring, Decompression, and Denoising.}
    For deblurring, decompression, and denoising tasks, we reuse the aforementioned general-purpose real-world video super-resolution models (SeedVR2-3B, FlashVSR, DOVE).
    Empirically, these diffusion-based models demonstrate strong robustness to multi-order degradations such as blur, compression artifacts, and noise, outperforming task-specific counterparts in complex real-world scenarios.

\paragraph{Frame Interpolation.} GIMM-VFI~\citep{guo2024generalizable} is a flow-based video frame interpolation method that models spatiotemporal motion using an implicit representation. It encodes bidirectional optical flows into a motion latent and predicts arbitrary-timestep flows via a coordinate-based neural network conditioned on spatial--temporal coordinates.
In our setting, we find that this operator is sufficient to handle most cases and achieves consistently strong performance.

\subsection{Prompts of MLLM Experts}
We have presented the prompts used for MLLM expert comparison in Tab.~\ref{tab:prompt-appendix}.

\begin{table*}[t!]
    \centering
    \renewcommand{\arraystretch}{1.5}
    \caption{Prompts for the MLLM comparison experts and for the VQ-Jarvis.}
    \label{tab:prompt-appendix}
    \begin{tabular}{p{13.5cm}}
     \shline
     \textbf{Prompt for MLLM Expert:} You will see two videos represented by sampled frames. Treat the first as Video A and the second as Video B. Decide which is better. Output exactly: Answer: $<$Video A, Video B, Tie$>$.\\
     \hline
     \textbf{Prompt for Multiple Order Judge Model:} Video A and Video B are results from two video enhancement methods. Explain and justify whether Video A or Video B is better. You should think from the following aspects: (1) Visual quality, (2) temporal consistency, (3) dynamic degree. \\
     \hline
     \textbf{Prompt for Video Scoring:} Please rate the quality of this video with a score accurate to two decimal places. The range of quality score is between 1-5. Explain the reasoning process concisely from spatial quality, temporal quality, and dynamic degree. \\
     \hline
    \textbf{Prompt for Degradation Detection:}  Detect whether each degradation exists in this video. Return the result in $<$answer$>$...$<$/answer$>$ as JSON with keys: ``dark", ``rain", ``blur", ``compression", ``noise", ``low-resolution", ``low-frame". \\
     \shline
    \end{tabular}
\end{table*}

\begin{figure*}[t!]
	\centering
    \includegraphics[width=1.0\linewidth]{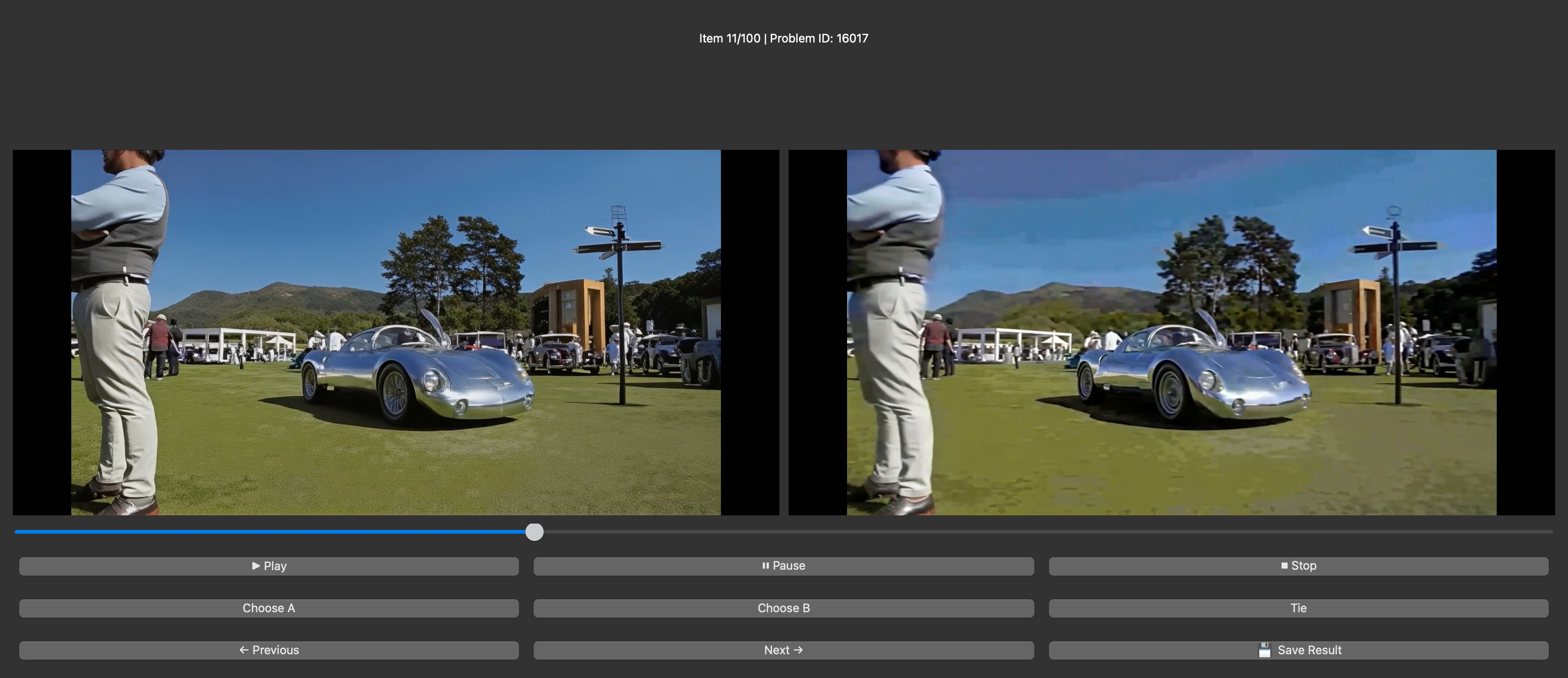}
	\vspace{-10pt}
	\caption{GUI used for human annotations.}
	\label{gui}
    \vspace{-10pt}
\end{figure*} 

\subsection{GUI of Human Annotations}
Fig.~\ref{gui} illustrates the graphical user interface used for video comparison annotation. The interface presents two videos side by side and supports synchronized playback control, allowing annotators to directly compare visual quality, temporal consistency, and overall perceptual preference. Annotators are required to select one of three options, \emph{A}, \emph{B}, or \emph{Tie}, to indicate which video is preferred or whether no clear preference exists.

We recruit 10 volunteers to perform the annotation using this interface. During annotation, the left and right videos are played simultaneously to facilitate direct comparison. Each video pair is independently annotated by three different volunteers for cross-validation. To assess intra-annotator consistency, a subset of video pairs is deliberately repeated at different stages of the annotation process, enabling us to verify the stability of volunteer judgments over time.

In addition, we randomly sample 1,000 video pairs from machine-generated comparison results and ask human annotators to verify whether the predicted preferences align with human perception. Furthermore, we invite human experts to annotate an additional 2,000 video pairs, providing high-quality preference labels to further enrich the dataset and improve the overall reliability of the preference annotations.

\subsection{Visualization of Training Samples on the VSR-Compare}
\begin{figure*}[t!]
	\centering
    \includegraphics[width=1.0\linewidth]{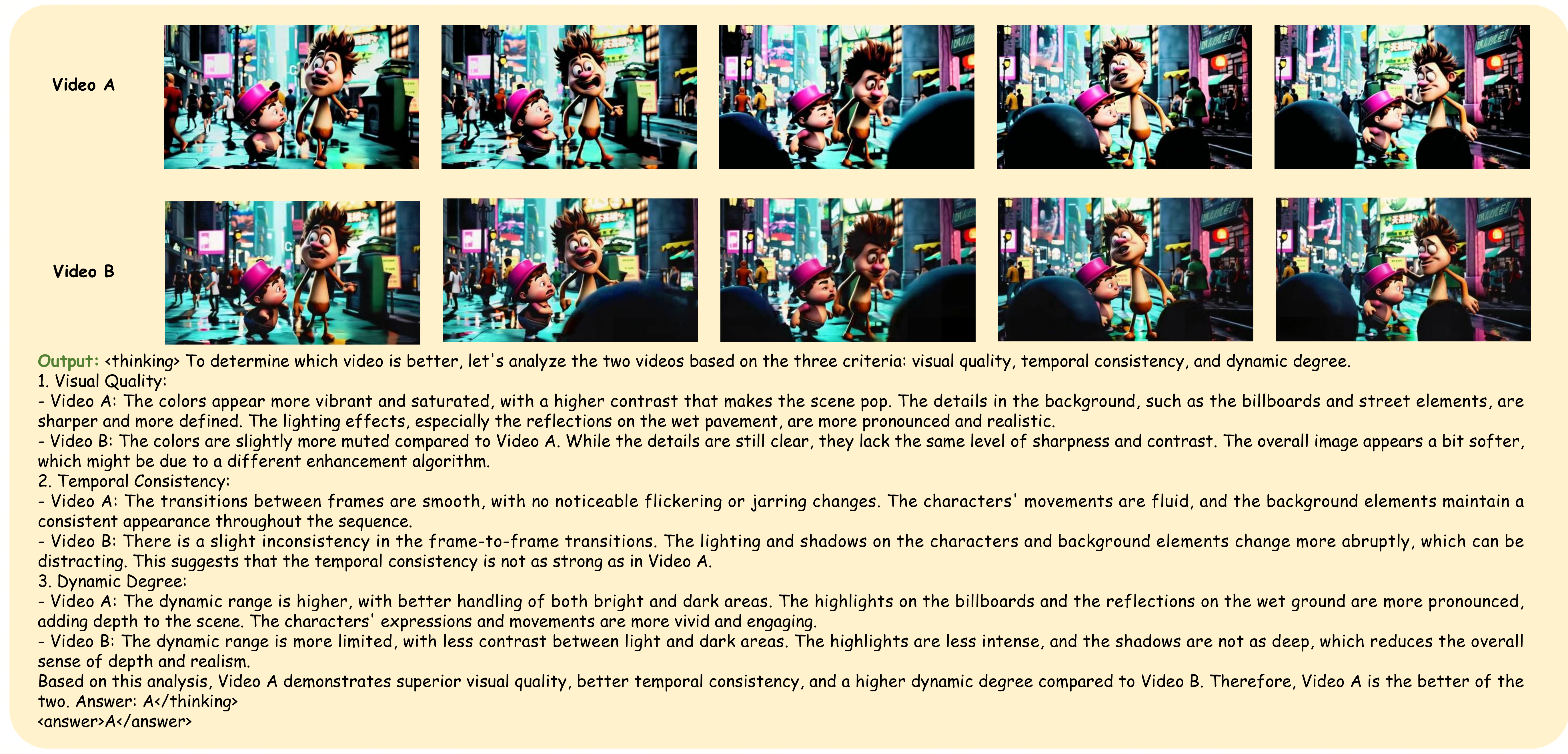}
	\vspace{-10pt}
	\caption{Low-light enhancement paired comparison training samples of our VSR-Compare.}
	\label{sample1}
    \vspace{-10pt}
\end{figure*}

\begin{figure*}[t!]
	\centering
    \includegraphics[width=1.0\linewidth]{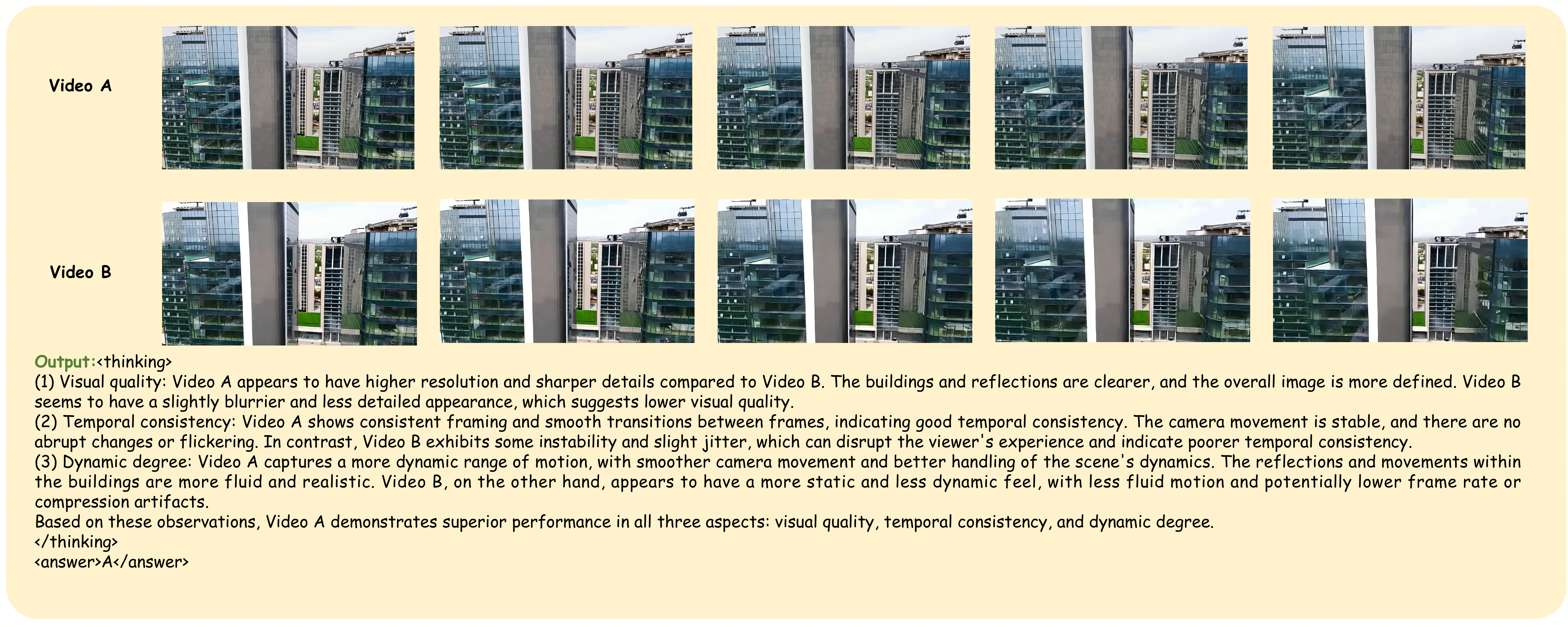}
	\vspace{-10pt}
	\caption{Super-resolution paired comparison training samples of our VSR-Compare.}
	\label{sample2}
    \vspace{-10pt}
\end{figure*}

We present the visualized training samples of our VSR-Compare. Fig.~\ref{sample1} presents a low-light enhancement paired comparison case, where Video B shows almost no noticeable signs of brightness enhancement. Fig.~\ref{sample2} shows a super-resolution paired comparison case, where Video A is sharper with better temporal consistency. The reasoning output is fused and aligned from different MLLM experts.

\section{More Details of Judge and Degradation Model}

\subsection{Details of Training Degradation Perception and Multiple Operator Model}

We adopt Group Relative Policy Optimization (GRPO)~\cite{guo2025deepseek, xu2025avatarshield} as the core training framework for the degradation detection and multi-operator judge model of VQ-Jarvis, which is used to optimize the agent’s decision policy for restoration operator scheduling.
Unlike conventional reinforcement learning approaches such as PPO that rely on a dedicated value critic, GRPO learns from relative comparisons among multiple candidate decisions, making it particularly suitable for preference-based supervision and combinatorial decision problems.

In the context of VQ-Jarvis, each query $q$ corresponds to a degraded input video together with its predicted degradation attributes.
Given $q$, the agent samples a group of $N$ candidate samples
$\{o_1, o_2, \dots, o_N\}$ from the current or previous policy $\pi_{\theta_{\text{old}}}$, where each $o_i$ represents a complete operator sequence for multi-degradation restoration.
Each trajectory is executed to produce a reconstructed video, which is then evaluated by Eq.~\ref{eq:score}-~\ref{eq:comp} to obtain scalar rewards
$\{r_1, r_2, \ldots, r_N\}$.

Instead of estimating absolute value functions, GRPO computes a relative advantage for each trajectory by normalizing its reward with respect to the group statistics:
\begin{equation}
\hat{A}_{i} = \frac{r_i - \operatorname{mean}(\{r_1, r_2, \ldots, r_N\})}{\operatorname{std}(\{r_1, r_2, \ldots, r_N\})}.
\end{equation}
After computing the relative advantages, GRPO updates the policy by comparing the likelihood of each trajectory under the new policy $\pi_{\theta_{\text{new}}}$ and the old policy $\pi_{\theta_{\text{old}}}$.
To ensure stable training, the likelihood ratio is clipped within $[1-\delta, 1+\delta]$, and a KL divergence term is introduced to regularize the policy toward a reference model.
The final optimization objective is given by:
\begin{equation}
\begin{aligned}
\mathcal{J}_{\text{GRPO}}(\theta)
&= \mathbb{E}_{[q \sim Q,\, o_i \sim \pi_{\theta_{\text{old}}}]}
\Bigg\{ \\
&\quad \min \Big[
    \rho_i \hat{A}_i,\,
    \operatorname{clip}(\rho_i, 1-\delta, 1+\delta)\hat{A}_i
\Big] \\
&\quad - \beta \cdot \mathbb{D}_{\text{KL}}[\pi_\theta \| \pi_{\text{ref}}]
\Bigg\}
\end{aligned}
\end{equation}
where $\rho_i = \pi_{\theta_{\text{new}}}(o_i \mid q)/\pi_{\theta_{\text{old}}}(o_i \mid q)$ denotes the policy update ratio, $\delta$ controls update stability, and $\beta$ balances the KL regularization.
$Q$ denotes the set of training video queries.

Qwen-3-VL-8B-Instruct~\cite{yang2025qwen3} is used as our pretrained VLM. In the GRPO algorithm, the generation number $N$ is set to $8$, the weight of KL divergence penalty $\beta$ is set to $0.001$, while the weights $\alpha_1$ and $\alpha_2$ are set to $0.25$ and $0.75$, respectively. We employ AdamW~\cite{loshchilov2017decoupled} as the optimizer, using an initial learning rate of $1\times10^{-6}$ that linearly decays to $1\times 10^{-9}$ during training.


\subsection{Prompts of our VQ-Jarvis}
We have presented the prompts for $\mathcal{H}_{\text{deg}}$ and $\mathcal{H}_{\text{judge}}$ in Tab.~\ref{tab:prompt-appendix}.
\clearpage

\end{appendices}
\clearpage
\bibliography{sn-bibliography}

@inproceedings{wu2023exploring,
  title={Exploring video quality assessment on user generated contents from aesthetic and technical perspectives},
  author={Wu, Haoning and Zhang, Erli and Liao, Liang and Chen, Chaofeng and Hou, Jingwen and Wang, Annan and Sun, Wenxiu and Yan, Qiong and Lin, Weisi},
  booktitle={Proceedings of the IEEE/CVF International Conference on Computer Vision},
  pages={20144--20154},
  year={2023}
}

@article{loshchilov2017decoupled,
  title={Decoupled weight decay regularization},
  author={Loshchilov, Ilya and Hutter, Frank},
  journal={arXiv preprint arXiv:1711.05101},
  year={2017}
}

@inproceedings{wang2023benchmark,
  title={Benchmark dataset and effective inter-frame alignment for real-world video super-resolution},
  author={Wang, Ruohao and Liu, Xiaohui and Zhang, Zhilu and Wu, Xiaohe and Feng, Chun-Mei and Zhang, Lei and Zuo, Wangmeng},
  booktitle={Proceedings of the IEEE/CVF conference on computer vision and pattern recognition},
  pages={1168--1177},
  year={2023}
}

@article{liang2022recurrent,
  title={Recurrent video restoration transformer with guided deformable attention},
  author={Liang, Jingyun and Fan, Yuchen and Xiang, Xiaoyu and Ranjan, Rakesh and Ilg, Eddy and Green, Simon and Cao, Jiezhang and Zhang, Kai and Timofte, Radu and Gool, Luc V},
  journal={Advances in Neural Information Processing Systems},
  volume={35},
  pages={378--393},
  year={2022}
}

@article{shi2022rethinking,
  title={Rethinking alignment in video super-resolution transformers},
  author={Shi, Shuwei and Gu, Jinjin and Xie, Liangbin and Wang, Xintao and Yang, Yujiu and Dong, Chao},
  journal={Advances in Neural Information Processing Systems},
  volume={35},
  pages={36081--36093},
  year={2022}
}

@inproceedings{chen2025dove,
  title={DOVE: Efficient One-Step Diffusion Model for Real-World Video Super-Resolution},
  author={Chen, Zheng and Zou, Zichen and Zhang, Kewei and Su, Xiongfei and Yuan, Xin and Guo, Yong and Zhang, Yulun},
  booktitle={Advances in Neural Information Processing Systems},
  year={2025}
}

@inproceedings{wang2025seedvr2,
  title={Seedvr2: One-step video restoration via diffusion adversarial post-training},
  author={Wang, Jianyi and Lin, Shanchuan and Lin, Zhijie and Ren, Yuxi and Wei, Meng and Yue, Zongsheng and Zhou, Shangchen and Chen, Hao and Zhao, Yang and Yang, Ceyuan and others},
  booktitle={International Conference on Learning Representations},
  year={2026}
}

@inproceedings{tao2017detail,
  title={Detail-revealing deep video super-resolution},
  author={Tao, Xin and Gao, Hongyun and Liao, Renjie and Wang, Jue and Jia, Jiaya},
  booktitle={Proceedings of the IEEE international conference on computer vision},
  pages={4472--4480},
  year={2017}
}

@inproceedings{nah2019ntire,
  title={Ntire 2019 challenge on video deblurring and super-resolution: Dataset and study},
  author={Nah, Seungjun and Baik, Sungyong and Hong, Seokil and Moon, Gyeongsik and Son, Sanghyun and Timofte, Radu and Mu Lee, Kyoung},
  booktitle={Proceedings of the IEEE/CVF conference on computer vision and pattern recognition workshops},
  year={2019}
}

@inproceedings{wang2025seedvr,
  title={Seedvr: Seeding infinity in diffusion transformer towards generic video restoration},
  author={Wang, Jianyi and Lin, Zhijie and Wei, Meng and Zhao, Yang and Yang, Ceyuan and Loy, Chen Change and Jiang, Lu},
  booktitle={Proceedings of the Computer Vision and Pattern Recognition Conference},
  pages={2161--2172},
  year={2025}
}

@inproceedings{yang2021real,
  title={Real-world video super-resolution: A benchmark dataset and a decomposition based learning scheme},
  author={Yang, Xi and Xiang, Wangmeng and Zeng, Hui and Zhang, Lei},
  booktitle={Proceedings of the IEEE/CVF international conference on computer vision},
  pages={4781--4790},
  year={2021}
}

@article{he2024venhancer,
  title={Venhancer: Generative space-time enhancement for video generation},
  author={He, Jingwen and Xue, Tianfan and Liu, Dongyang and Lin, Xinqi and Gao, Peng and Lin, Dahua and Qiao, Yu and Ouyang, Wanli and Liu, Ziwei},
  journal={arXiv preprint arXiv:2407.07667},
  year={2024}
}

@article{xie2025star,
  title={Star: Spatial-temporal augmentation with text-to-video models for real-world video super-resolution},
  author={Xie, Rui and Liu, Yinhong and Zhou, Penghao and Zhao, Chen and Zhou, Jun and Zhang, Kai and Zhang, Zhenyu and Yang, Jian and Yang, Zhenheng and Tai, Ying},
  journal={arXiv preprint arXiv:2501.02976},
  year={2025}
}

@inproceedings{yang2024motion,
  title={Motion-guided latent diffusion for temporally consistent real-world video super-resolution},
  author={Yang, Xi and He, Chenhang and Ma, Jianqi and Zhang, Lei},
  booktitle={European conference on computer vision},
  pages={224--242},
  year={2024},
  organization={Springer}
}

@inproceedings{zhou2024upscale,
  title={Upscale-a-video: Temporal-consistent diffusion model for real-world video super-resolution},
  author={Zhou, Shangchen and Yang, Peiqing and Wang, Jianyi and Luo, Yihang and Loy, Chen Change},
  booktitle={Proceedings of the IEEE/CVF Conference on Computer Vision and Pattern Recognition},
  pages={2535--2545},
  year={2024}
}

@article{liang2024vrt,
  title={Vrt: A video restoration transformer},
  author={Liang, Jingyun and Cao, Jiezhang and Fan, Yuchen and Zhang, Kai and Ranjan, Rakesh and Li, Yawei and Timofte, Radu and Van Gool, Luc},
  journal={IEEE Transactions on Image Processing},
  volume={33},
  pages={2171--2182},
  year={2024},
  publisher={IEEE}
}

@inproceedings{wang2019edvr,
  title={Edvr: Video restoration with enhanced deformable convolutional networks},
  author={Wang, Xintao and Chan, Kelvin CK and Yu, Ke and Dong, Chao and Loy, Chen Change},
  booktitle={Proceedings of the IEEE/CVF conference on computer vision and pattern recognition workshops},
  year={2019}
}

@inproceedings{chan2022basicvsr++,
  title={Basicvsr++: Improving video super-resolution with enhanced propagation and alignment},
  author={Chan, Kelvin CK and Zhou, Shangchen and Xu, Xiangyu and Loy, Chen Change},
  booktitle={Proceedings of the IEEE/CVF conference on computer vision and pattern recognition},
  pages={5972--5981},
  year={2022}
}

@inproceedings{chan2021basicvsr,
  title={Basicvsr: The search for essential components in video super-resolution and beyond},
  author={Chan, Kelvin CK and Wang, Xintao and Yu, Ke and Dong, Chao and Loy, Chen Change},
  booktitle={Proceedings of the IEEE/CVF conference on computer vision and pattern recognition},
  pages={4947--4956},
  year={2021}
}

@inproceedings{chan2022investigating,
  title={Investigating tradeoffs in real-world video super-resolution},
  author={Chan, Kelvin CK and Zhou, Shangchen and Xu, Xiangyu and Loy, Chen Change},
  booktitle={Proceedings of the IEEE/CVF conference on computer vision and pattern recognition},
  pages={5962--5971},
  year={2022}
}

@inproceedings{yang2022maniqa,
  title={Maniqa: Multi-dimension attention network for no-reference image quality assessment},
  author={Yang, Sidi and Wu, Tianhe and Shi, Shuwei and Lao, Shanshan and Gong, Yuan and Cao, Mingdeng and Wang, Jiahao and Yang, Yujiu},
  booktitle={Proceedings of the IEEE/CVF Conference on Computer Vision and Pattern Recognition},
  pages={1191--1200},
  year={2022}
}

@inproceedings{wang2021real,
  title={Real-ESRGAN: Training real-world blind super-resolution with pure synthetic data},
  author={Wang, Xintao and Xie, Liangbin and Dong, Chao and Shan, Ying},
  booktitle={Proceedings of the IEEE/CVF International Conference on Computer Vision},
  pages={1905--1914},
  year={2021}
}

@inproceedings{zhang2021designing,
  title={Designing a practical degradation model for deep blind image super-resolution},
  author={Zhang, Kai and Liang, Jingyun and Van Gool, Luc and Timofte, Radu},
  booktitle={Proceedings of the IEEE/CVF International Conference on Computer Vision},
  pages={4791--4800},
  year={2021}
}

@article{zhuang2025flashvsr,
  title={FlashVSR: Towards Real-Time Diffusion-Based Streaming Video Super-Resolution},
  author={Zhuang, Junhao and Guo, Shi and Cai, Xin and Li, Xiaohui and Liu, Yihao and Yuan, Chun and Xue, Tianfan},
  journal={arXiv preprint arXiv:2510.12747},
  year={2025}
}

@inproceedings{li2025q,
  title={Q-insight: Understanding image quality via visual reinforcement learning},
  author={Li, Weiqi and Zhang, Xuanyu and Zhao, Shijie and Zhang, Yabin and Li, Junlin and Zhang, Li and Zhang, Jian},
  booktitle={Advances in Neural Information Processing Systems},
  year={2025}
}

@inproceedings{wang2023exploring,
  title={Exploring clip for assessing the look and feel of images},
  author={Wang, Jianyi and Chan, Kelvin CK and Loy, Chen Change},
  booktitle={Proceedings of the AAAI conference on artificial intelligence},
  volume={37},
  number={2},
  pages={2555--2563},
  year={2023}
}

@inproceedings{ying2021patch,
  title={Patch-vq:'patching up'the video quality problem},
  author={Ying, Zhenqiang and Mandal, Maniratnam and Ghadiyaram, Deepti and Bovik, Alan},
  booktitle={Proceedings of the IEEE/CVF conference on computer vision and pattern recognition},
  pages={14019--14029},
  year={2021}
}

@article{sinno2018large,
  title={Large-scale study of perceptual video quality},
  author={Sinno, Zeina and Bovik, Alan Conrad},
  journal={IEEE Transactions on Image Processing},
  volume={28},
  number={2},
  pages={612--627},
  year={2018},
  publisher={IEEE}
}

@inproceedings{hosu2017konstanz,
  title={The Konstanz natural video database (KoNViD-1k)},
  author={Hosu, Vlad and Hahn, Franz and Jenadeleh, Mohsen and Lin, Hanhe and Men, Hui and Szir{\'a}nyi, Tam{\'a}s and Li, Shujun and Saupe, Dietmar},
  booktitle={2017 Ninth international conference on quality of multimedia experience (QoMEX)},
  pages={1--6},
  year={2017},
  organization={IEEE}
}

@article{xu2025avatarshield,
  title={AvatarShield: Visual Reinforcement Learning for Human-Centric Video Forgery Detection},
  author={Xu, Zhipei and Zhang, Xuanyu and Zhou, Xing and Zhang, Jian},
  journal={arXiv preprint arXiv:2505.15173},
  year={2025}
}

@article{guo2025deepseek,
  title={Deepseek-r1: Incentivizing reasoning capability in llms via reinforcement learning},
  author={Guo, Daya and Yang, Dejian and Zhang, Haowei and Song, Junxiao and Zhang, Ruoyu and Xu, Runxin and Zhu, Qihao and Ma, Shirong and Wang, Peiyi and Bi, Xiao and others},
  journal={arXiv preprint arXiv:2501.12948},
  year={2025}
}

@inproceedings{wu2022fast,
  title={Fast-vqa: Efficient end-to-end video quality assessment with fragment sampling},
  author={Wu, Haoning and Chen, Chaofeng and Hou, Jingwen and Liao, Liang and Wang, Annan and Sun, Wenxiu and Yan, Qiong and Lin, Weisi},
  booktitle={European conference on computer vision},
  pages={538--554},
  year={2022},
  organization={Springer}
}

@inproceedings{wu2024q,
  title={Q-ALIGN: teaching LMMs for visual scoring via discrete text-defined levels},
  author={Wu, Haoning and Zhang, Zicheng and Zhang, Weixia and Chen, Chaofeng and Liao, Liang and Li, Chunyi and Gao, Yixuan and Wang, Annan and Zhang, Erli and Sun, Wenxiu and others},
  booktitle={Proceedings of the 41st International Conference on Machine Learning},
  pages={54015--54029},
  year={2024}
}

@article{jia2024vqa,
  title={VQA2: Visual Question Answering for Video Quality Assessment},
  author={Jia, Ziheng and Zhang, Zicheng and Qian, Jiaying and Wu, Haoning and Sun, Wei and Li, Chunyi and Liu, Xiaohong and Lin, Weisi and Zhai, Guangtao and Min, Xiongkuo},
  journal={arXiv preprint arXiv:2411.03795},
  year={2024}
}

@inproceedings{wu2024qinstruct,
  title={Q-instruct: Improving low-level visual abilities for multi-modality foundation models},
  author={Wu, Haoning and Zhang, Zicheng and Zhang, Erli and Chen, Chaofeng and Liao, Liang and Wang, Annan and Xu, Kaixin and Li, Chunyi and Hou, Jingwen and Zhai, Guangtao and others},
  booktitle={Proceedings of the IEEE/CVF conference on computer vision and pattern recognition},
  pages={25490--25500},
  year={2024}
}

@article{sun2024analysis,
  title={Analysis of video quality datasets via design of minimalistic video quality models},
  author={Sun, Wei and Wen, Wen and Min, Xiongkuo and Lan, Long and Zhai, Guangtao and Ma, Kede},
  journal={IEEE Transactions on Pattern Analysis and Machine Intelligence},
  year={2024}
}

@article{yang2024cogvideox,
  title={Cogvideox: Text-to-video diffusion models with an expert transformer},
  author={Yang, Zhuoyi and Teng, Jiayan and Zheng, Wendi and Ding, Ming and Huang, Shiyu and Xu, Jiazheng and Yang, Yuanming and Hong, Wenyi and Zhang, Xiaohan and Feng, Guanyu and others},
  journal={arXiv preprint arXiv:2408.06072},
  year={2024}
}

@inproceedings{zhao2025reasoning,
  title={Reasoning as Representation: Rethinking Visual Reinforcement Learning in Image Quality Assessment},
  author={Zhao, Shijie and Zhang, Xuanyu and Li, Weiqi and Li, Junlin and Zhang, Li and Xue, Tianfan and Zhang, Jian},
  booktitle={International Conference on Learning Representations},
  year={2026}
}

@article{liu2025moa,
  title={MoA-VR: A Mixture-of-Agents System Towards All-in-One Video Restoration},
  author={Liu, Lu and Cai, Chunlei and Shen, Shaocheng and Liang, Jianfeng and Ouyang, Weimin and Ye, Tianxiao and Mao, Jian and Duan, Huiyu and Yao, Jiangchao and Zhang, Xiaoyun and others},
  journal={arXiv preprint arXiv:2510.08508},
  year={2025}
}

@article{potlapalli2023promptir,
  title={Promptir: Prompting for all-in-one image restoration},
  author={Potlapalli, Vaishnav and Zamir, Syed Waqas and Khan, Salman H and Shahbaz Khan, Fahad},
  journal={Advances in Neural Information Processing Systems},
  volume={36},
  pages={71275--71293},
  year={2023}
}

@inproceedings{wang2023reparo,
  title={Reparo: QoE-aware live video streaming in low-rate networks by intelligent frame recovery},
  author={Wang, Fulin and Li, Qing and Shi, Wanxin and Tyson, Gareth and Jiang, Yong and Ma, Lianbo and Zhang, Peng and Lan, Yulong and Li, Zhicheng},
  booktitle={Proceedings of the 31st ACM International Conference on Multimedia},
  pages={9194--9204},
  year={2023}
}

@inproceedings{lin2025jarvisir,
  title={Jarvisir: Elevating autonomous driving perception with intelligent image restoration},
  author={Lin, Yunlong and Lin, Zixu and Chen, Haoyu and Pan, Panwang and Li, Chenxin and Chen, Sixiang and Wen, Kairun and Jin, Yeying and Li, Wenbo and Ding, Xinghao},
  booktitle={Proceedings of the Computer Vision and Pattern Recognition Conference},
  pages={22369--22380},
  year={2025}
}

@inproceedings{zhu2024intelligent,
  title={An intelligent agentic system for complex image restoration problems},
  author={Zhu, Kaiwen and Gu, Jinjin and You, Zhiyuan and Qiao, Yu and Dong, Chao},
booktitle={International Conference on Learning Representations},
  year={2024}
}

@article{jiang2025multi,
  title={Multi-Agent Image Restoration},
  author={Jiang, Xu and Li, Gehui and Chen, Bin and Zhang, Jian},
  journal={arXiv preprint arXiv:2503.09403},
  year={2025}
}

@article{zhou2025q,
  title={Q-Agent: Quality-Driven Chain-of-Thought Image Restoration Agent through Robust Multimodal Large Language Model},
  author={Zhou, Yingjie and Cao, Jiezhang and Zhang, Zicheng and Wen, Farong and Jiang, Yanwei and Jia, Jun and Liu, Xiaohong and Min, Xiongkuo and Zhai, Guangtao},
  journal={arXiv preprint arXiv:2504.07148},
  year={2025}
}

@inproceedings{zuo20254kagent,
  title={4kagent: agentic any image to 4k super-resolution},
  author={Zuo, Yushen and Zheng, Qi and Wu, Mingyang and Jiang, Xinrui and Li, Renjie and Wang, Jian and Zhang, Yide and Mai, Gengchen and Wang, Lihong V and Zou, James and others},
  booktitle={Advances in Neural Information Processing Systems},
  year={2025}
}

@inproceedings{chen2024restoreagent,
  title={Restoreagent: Autonomous image restoration agent via multimodal large language models},
  author={Chen, Haoyu and Li, Wenbo and Gu, Jinjin and Ren, Jingjing and Chen, Sixiang and Ye, Tian and Pei, Renjing and Zhou, Kaiwen and Song, Fenglong and Zhu, Lei},
  booktitle={Advances in Neural Information Processing Systems},
  volume={37},
  pages={110643--110666},
  year={2024}
}

@article{li2025hybrid,
  title={Hybrid agents for image restoration},
  author={Li, Bingchen and Li, Xin and Lu, Yiting and Chen, Zhibo},
  journal={arXiv preprint arXiv:2503.10120},
  year={2025}
}

@inproceedings{li2025codetree,
  title={Codetree: Agent-guided tree search for code generation with large language models},
  author={Li, Jierui and Le, Hung and Zhou, Yingbo and Xiong, Caiming and Savarese, Silvio and Sahoo, Doyen},
  booktitle={Proceedings of the 2025 Conference of the Nations of the Americas Chapter of the Association for Computational Linguistics: Human Language Technologies (Volume 1: Long Papers)},
  pages={3711--3726},
  year={2025}
}

@inproceedings{wen2024modular,
  title={Modular blind video quality assessment},
  author={Wen, Wen and Li, Mu and Zhang, Yabin and Liao, Yiting and Li, Junlin and Zhang, Li and Ma, Kede},
  booktitle={Proceedings of the IEEE/CVF Conference on Computer Vision and Pattern Recognition},
  pages={2763--2772},
  year={2024}
}

@inproceedings{zhang2025vq,
  title={VQ-Insight: Teaching VLMs for AI-Generated Video Quality Understanding via Progressive Visual Reinforcement Learning},
  author={Zhang, Xuanyu and Li, Weiqi and Zhao, Shijie and Li, Junlin and Zhang, Li and Zhang, Jian},
  booktitle={Proceedings of the AAAI Conference on Artificial Intelligence},
  year={2026}
}

@inproceedings{chen2025toward,
  title={Toward generalized image quality assessment: Relaxing the perfect reference quality assumption},
  author={Chen, Du and Wu, Tianhe and Ma, Kede and Zhang, Lei},
  booktitle={Proceedings of the Computer Vision and Pattern Recognition Conference},
  pages={12742--12752},
  year={2025}
}

@inproceedings{chen2024comparative,
  title={A comparative study of image restoration networks for general backbone network design},
  author={Chen, Xiangyu and Li, Zheyuan and Pu, Yuandong and Liu, Yihao and Zhou, Jiantao and Qiao, Yu and Dong, Chao},
  booktitle={European Conference on Computer Vision},
  pages={74--91},
  year={2024},
  organization={Springer}
}

@article{zheng2024lm4lv,
  title={Lm4lv: A frozen large language model for low-level vision tasks},
  author={Zheng, Boyang and Gu, Jinjin and Li, Shijun and Dong, Chao},
  journal={arXiv preprint arXiv:2405.15734},
  year={2024}
}

@article{li2025investigate,
  title={Investigate the Low-level Visual Perception in Vision-Language based Image Quality Assessment},
  author={Li, Yuan and Sun, Zitang and Chen, Yen-Ju and Nishida, Shin'ya},
  journal={arXiv preprint arXiv:2512.09573},
  year={2025}
}

@inproceedings{wu2024rainmamba,
  title={Rainmamba: Enhanced locality learning with state space models for video deraining},
  author={Wu, Hongtao and Yang, Yijun and Xu, Huihui and Wang, Weiming and Zhou, Jinni and Zhu, Lei},
  booktitle={Proceedings of the 32nd ACM International Conference on Multimedia},
  pages={7881--7890},
  year={2024}
}

@inproceedings{conde2024instructir,
  title={Instructir: High-quality image restoration following human instructions},
  author={Conde, Marcos V and Geigle, Gregor and Timofte, Radu},
  booktitle={European Conference on Computer Vision},
  pages={1--21},
  year={2024},
  organization={Springer}
}

@article{guo2024generalizable,
  title={Generalizable implicit motion modeling for video frame interpolation},
  author={Guo, Zujin and Li, Wei and Loy, Chen Change},
  journal={Advances in Neural Information Processing Systems},
  volume={37},
  pages={63747--63770},
  year={2024}
}

@inproceedings{duan2025finevq,
  title={Finevq: Fine-grained user generated content video quality assessment},
  author={Duan, Huiyu and Hu, Qiang and Wang, Jiarui and Yang, Liu and Xu, Zitong and Liu, Lu and Min, Xiongkuo and Cai, Chunlei and Ye, Tianxiao and Zhang, Xiaoyun and others},
  booktitle={Proceedings of the Computer Vision and Pattern Recognition Conference},
  pages={3206--3217},
  year={2025}
}

@inproceedings{sun2022deep,
  title={A deep learning based no-reference quality assessment model for ugc videos},
  author={Sun, Wei and Min, Xiongkuo and Lu, Wei and Zhai, Guangtao},
  booktitle={Proceedings of the 30th ACM International Conference on Multimedia},
  pages={856--865},
  year={2022}
}

@inproceedings{li2019quality,
  title={Quality assessment of in-the-wild videos},
  author={Li, Dingquan and Jiang, Tingting and Jiang, Ming},
  booktitle={Proceedings of the 27th ACM international conference on multimedia},
  pages={2351--2359},
  year={2019}
}

@article{chen2021learning,
  title={Learning generalized spatial-temporal deep feature representation for no-reference video quality assessment},
  author={Chen, Baoliang and Zhu, Lingyu and Li, Guo and Lu, Fangbo and Fan, Hongfei and Wang, Shiqi},
  journal={IEEE Transactions on Circuits and Systems for Video Technology},
  volume={32},
  number={4},
  pages={1903--1916},
  year={2021},
  publisher={IEEE}
}

@article{mittal2012making,
  title={Making a “completely blind” image quality analyzer},
  author={Mittal, Anish and Soundararajan, Rajiv and Bovik, Alan C},
  journal={IEEE Signal processing letters},
  volume={20},
  number={3},
  pages={209--212},
  year={2012},
  publisher={IEEE}
}

@article{tu2021ugc,
  title={UGC-VQA: Benchmarking blind video quality assessment for user generated content},
  author={Tu, Zhengzhong and Wang, Yilin and Birkbeck, Neil and Adsumilli, Balu and Bovik, Alan C},
  journal={IEEE Transactions on Image Processing},
  volume={30},
  pages={4449--4464},
  year={2021},
  publisher={IEEE}
}

@article{tu2021rapique,
  title={RAPIQUE: Rapid and accurate video quality prediction of user generated content},
  author={Tu, Zhengzhong and Yu, Xiangxu and Wang, Yilin and Birkbeck, Neil and Adsumilli, Balu and Bovik, Alan C},
  journal={IEEE Open Journal of Signal Processing},
  volume={2},
  pages={425--440},
  year={2021},
  publisher={IEEE}
}

@article{yang2025qwen3,
  title={Qwen3 technical report},
  author={Yang, An and Li, Anfeng and Yang, Baosong and Zhang, Beichen and Hui, Binyuan and Zheng, Bo and Yu, Bowen and Gao, Chang and Huang, Chengen and Lv, Chenxu and others},
  journal={arXiv preprint arXiv:2505.09388},
  year={2025}
}

@article{bai2025qwen3vl,
  title={Qwen3-VL technical report},
  author={Bai, Shuai and Cai, Yuxuan and Chen, Ruizhe and Chen, Keqin and Chen, Xionghui and Cheng, Zesen and others},
  journal={arXiv preprint arXiv:2511.21631},
  year={2025}
}

@article{li2026qwen3,
  title={Qwen3-VL-Embedding and Qwen3-VL-Reranker: A Unified Framework for State-of-the-Art Multimodal Retrieval and Ranking},
  author={Li, Mingxin and Zhang, Yanzhao and Long, Dingkun and Chen, Keqin and Song, Sibo and Bai, Shuai and Yang, Zhibo and Xie, Pengjun and Yang, An and Liu, Dayiheng and others},
  journal={arXiv preprint arXiv:2601.04720},
  year={2026}
}

@article{yang2025difflle,
  title={Difflle: Diffusion-based domain calibration for weak supervised low-light image enhancement},
  author={Yang, Shuzhou and Zhang, Xuanyu and Wang, Yinhuai and Yu, Jiwen and Wang, Yuhan and Zhang, Jian},
  journal={International Journal of Computer Vision},
  volume={133},
  number={5},
  pages={2527--2546},
  year={2025},
  publisher={Springer}
}

@article{peng2022lve,
  title={LVE-S2D: Low-light video enhancement from static to dynamic},
  author={Peng, Bo and Zhang, Xuanyu and Lei, Jianjun and Zhang, Zhe and Ling, Nam and Huang, Qingming},
  journal={IEEE Transactions on Circuits and Systems for Video Technology},
  volume={32},
  number={12},
  pages={8342--8352},
  year={2022},
  publisher={IEEE}
}

@inproceedings{
chen2026improved,
title={Improved Adversarial Diffusion Compression for Real-World Video Super-Resolution},
author={Bin Chen and Weiqi Li and Shijie Zhao and Xuanyu Zhang and Junlin Li and Li zhang and Jian Zhang},
booktitle={The Fourteenth International Conference on Learning Representations},
year={2026},
url={https://openreview.net/forum?id=U2SJE6W3wT}
}

@inproceedings{
wang2026gendr,
title={Gen{DR}: Lighten Generative Detail Restoration},
author={Yan Wang and Shijie Zhao and Kexin Zhang and Junlin Li and Li zhang},
booktitle={The Fourteenth International Conference on Learning Representations},
year={2026},
url={https://openreview.net/forum?id=vznIYSnv9J}
}

@inproceedings{li2025uare,
  title={UARE: A Unified Vision-Language Model for Image Quality Assessment, Restoration, and Enhancement},
  author={Li, Weiqi and Zhang, Xuanyu and Chen, Bin and Xie, Jingfen and Wang, Yan and Zhang, Kexin and Li, Junlin and Zhang, Li and Zhang, Jian and Zhao, Shijie},
  booktitle={Proceedings of the IEEE/CVF International Conference on Computer Vision},
  year={2026}
}

\end{document}